\documentclass{article}
\PassOptionsToPackage{numbers, compress}{natbib}

\usepackage[main, final]{neurips_2026}

\usepackage[utf8]{inputenc} %
\usepackage[T1]{fontenc}    %
\usepackage[hidelinks]{hyperref}       %
\usepackage{url}            %
\usepackage{booktabs}       %
\usepackage{amsfonts}       %
\usepackage{nicefrac}       %
\usepackage{microtype}      %
\usepackage{xcolor}         %

\usepackage{amsmath}
\usepackage{amssymb}
\usepackage{mathtools}
\usepackage{amsthm}
\usepackage{graphicx}
\usepackage{array}
\usepackage{multirow}
\usepackage[table]{xcolor}
\usepackage{subcaption}
\usepackage{rotating}
\usepackage{pdflscape}

\def\*#1{\mathbf{\bm #1}}
\usepackage{enumitem}

\newcommand{\B}{\mathcal B}

\newcommand{\X}{\mathcal X}

\usepackage[dvipsnames]{xcolor}

\usepackage[normalem]{ulem}

\makeatletter
\long\def\flatfntext#1{\noindent$\m@th^{\@thefnmark}$#1}
\let\@makefntext\flatfntext
\def\flatfnsetup{\let\@makefntext\flatfntext}
\def\@thanks{\protect\flatfnsetup}
\makeatother

\title{Breakeven complexity: A new perspective\\on neural partial differential equation solvers}

\author{%
  \begin{minipage}[t]{0.45\textwidth}
    \centering
    \textbf{Yijing Zhang} \\
    {\normalfont University of Wisconsin--Madison} \\
    \texttt{yzhang2637@wisc.edu}
  \end{minipage}
  \hfill
  \begin{minipage}[t]{0.45\textwidth}
    \centering
    \textbf{Nicholas Roberts\thanks{Work completed while at the University of Wisconsin--Madison.}} \\
    {\normalfont Princeton University} \\
    \texttt{nick11roberts@princeton.edu}
  \end{minipage}
  \\[1.2cm]
  \begin{minipage}[t]{0.45\textwidth}
    \centering
    \textbf{Tanya Marwah\thanks{Work completed while at the Simons Foundation.}} \\
    Google DeepMind \\
    \texttt{tmarwah@google.com}
  \end{minipage}
  \hfill
  \begin{minipage}[t]{0.45\textwidth}
    \centering
    \textbf{Mikhail Khodak} \\
    University of Wisconsin--Madison \\
    \texttt{khodak@wisc.edu}
  \end{minipage}
}

\begin{document}

\maketitle

\begin{abstract}
Neural surrogate solvers of partial differential equations~(PDEs) promise dramatic speedups over numerical methods, especially in scenarios requiring many solves. 
However, current accuracy-based evaluations do not fully consider two central issues: (1)~neural solvers incur substantial up-front costs for data generation, training, and tuning; and (2)~classical solvers can also generate low-fidelity solutions at a sufficiently low simulation cost. 
To explicitly account for these realities and fully incorporate end-to-end costs, we propose an evaluation framework centered on {\bf breakeven complexity},\footnote{Code: \href{https://github.com/yijingz02/breakeven_complexity}{github.com/yijingz02/breakeven\_complexity}. Data: \href{https://huggingface.co/datasets/yijingz/breakeven_complexity}{huggingface.co/datasets/yijingz/breakeven\_complexity}.} a metric that counts the forward solves before a learned solver is cost-effective relative to an error-equivalent traditional solver.
To evaluate this measure, we apply scaling laws to determine how much training budget to allocate to data generation and discuss how to achieve smooth error-matching in diverse settings.
We evaluate the breakeven complexity of multiple  neural PDE solvers on three PDEs on 2D periodic domains from APEBench and a novel benchmark of flows past multiple obstacles generated by the GPU-native PyFR code.
Among other findings, our results suggest that neural PDE solvers become {\em more effective} as problems get harder in terms of cost, dimension, rollout, physics regime (e.g. higher Reynolds number), etc.\looseness-1
\end{abstract}

\vspace{-3mm}
\section{Introduction}
\vspace{-2mm}

PDE simulation is a crucial engineering tool that enables prediction and analysis of physics systems in settings where experiments are expensive or impractical~\citep{evans2022partial,leveque2007finite}.
Recently, advances in machine learning~(ML) have driven the proliferation of data-driven surrogate models that can quickly solve transient PDEs using only neural network forward passes~\cite{fno,walrus}.
This type of acceleration has significant potential impact across numerous large-scale scientific computing tasks.\looseness-1

However, while neural PDE surrogates often have fast inference when trained, they require substantial up-front cost, including data generation, model training and tuning. Furthermore, neural PDE surrogates are typically low-fidelity approximations, while classical simulators can trade-off accuracy and cost via tunable parameters (grid resolution, number of timesteps, etc.), thus also allowing for low-fidelity simulation at a reduced cost \cite{ashton2025fluid}. This creates a decision problem that accuracy alone does not address: 
when is it worthwhile to invest the up-front training cost of a surrogate, rather than simply running a numerical solver at an appropriately chosen lower fidelity?\looseness-1

In some settings, such as real-time inference, reducing latency is often the main concern, motivating surrogate deployment even with nontrivial training costs so long as the surrogate is faster than a classical solver of similar error~\citep{tangsali2025benchmarking, ashton2025fluid}.
We instead are motivated by cases where repeated forward PDE solves are required to fit latent quantities or optimize objectives, as in PDE-constrained inverse, control, and design problems~\citep{bochev2009optimization}.
For such non-real-time applications, the decision becomes fundamentally an amortization trade-off: 
will the surrogate solver be used enough times such that the compute spent on generating data and training the model is worth it, i.e. not better-spent on querying a traditional solver of similarly low fidelity?
Existing evaluation practice for neural PDE surrogates, which mostly focuses on measuring the predictive error against a fixed high-fidelity ground-truth simulator, typically under-represents end-to-end costs, making it difficult to answer this question~\citep{takamoto2022pdebench,koehler2024apebench,ohana2024well}.\looseness-1

To address this gap, we propose a compute-aware evaluation framework that makes the decision trade-off explicit.
Our approach is centered around \textbf{breakeven complexity}, which measures the number of forward solves required for a neural PDE surrogate to be cost-effective relative to an error-matched low-fidelity classical simulator.
Unlike accuracy-only evaluations, breakeven complexity accounts for data generation costs, training costs, inference costs, and low fidelity in a single number.
Reported jointly with error, breakeven complexity enables a more realistic comparison between learned models that better-informs decision-making and accurately conveys their potential relative to traditional solvers.\looseness-1

Our specific contributions are the following:\vspace{-2mm}
\begin{enumerate}[noitemsep, leftmargin=*]
    \item We build an evaluation framework around {\bf breakeven complexity}, a cost-aware metric for learned PDE solvers building on the amortization quantity suggested by \citet[Eq.~1]{weak_baseline}. We formulate its worst-case and average-case variants and make their computation practical by (a)~using scaling laws to allocate compute between optimization and data generation at a fixed budget level and (b)~showing how to smoothly vary the error while reducing resolution when finding an error-matched solver.\looseness-1
    \item We deploy our methodology on three 2D periodic PDE settings from APEBench, solved via the pseudo-spectral Exponax code~\citep{koehler2024apebench}, and on a novel dataset called {\bf BreakFlow} consisting of multi-obstacle flows simulated with the scale-resolving, GPU-native PyFR code~\citep{PyFR}.
    We release multi-fidelity variants of all four datasets along with protocols for enabling the evaluation of breakeven complexity on these and on future benchmarks.\looseness-1
    \item We evaluate the breakeven complexity of eight leading models (four supervised solvers and four foundation models) on the above benchmarks.
    Our investigation shows that models that do well according to reported error can sometimes underperform under our cost-aware metric, that it can be used to study the robustness of learned solvers, and that on the APEBench (periodic) tasks learned solvers can require hundreds of thousands of inference calls in order to be cost effective.\looseness-1
    \item On the other hand, by studying breakeven complexity across multiple spatial dimensions, rollout lengths, and Reynolds numbers, we find that it consistently {\em decreases} as the task gets harder.
    This suggests that the cost-effectiveness of neural PDE solvers may {\em improve} with simulation difficulty. \looseness-1
\end{enumerate}

\vspace{-3mm}
\section{Related Work}
\vspace{-2mm}

There are a wide variety of neural network-based approaches to solving PDEs, including physics-informed neural networks~(PINNs)~\citep{pinn} and neural operators~\citep{fno,ffno}.
Our work is centered around evaluating methods, such as the latter, which train networks on data generated by classical simulators;
this field has witnessed a steady improvement in (error-based) performance~\citep{ffno,eddyformer}.
Most recently, foundation models have begun to be used in this field as well, promising to alleviate the upfront cost of data generation via transfer from massive pretraining data~\citep{poseidon,walrus}.

These neural surrogates have been evaluated on a variety of benchmarks, largely via their predictive error relative to some high-fidelity numerical simulator.
Benchmarks such as PDEBench~\citep{takamoto2022pdebench}, PDEArena~\citep{gupta2022multispatiotemporalscalegeneralizedpdemodeling}, APEBench~\citep{koehler2024apebench} and the Well~\citep{ohana2024well} all provide large, ready-to-use datasets across different baseline PDEs with standardized metrics and baseline models. 
However, focusing on accuracy makes it hard to understand tradeoffs with cost, despite this being a crucial question in a field motivated largely by solver speed.
Our work addresses this issue by comparing explicitly to an error-matched traditional solver to derive a single interpretable value.\looseness-1

We are not the first to consider varying-fidelity classical solvers in evaluating learned approaches, which has been done for PINNs~\citep{sacchetti2022neural,grossmann2023physicsinformedneuralnetworksbeat} and (more relevantly) for neural operators by~\citet{weak_baseline}.
The latter showed the importance of simultaneously accounting for cost and accuracy and the need to compare to effective, well-suited, and problem-specific solvers;
these critiques directly motivate our work to develop a cost-aware basis for evaluating neural PDE solvers.
\citet{weak_baseline} also recommend the metric that in this work we make practical and call ``breakeven complexity'';
to our knowledge, the measure was not used by them nor in the subsequent literature.
\looseness-1

In addition to the framework and empirical evaluations, we also develop a new flow-past-multiple-obstacles benchmark called {\bf BreakFlow}.
While several flow-past-object tasks exist~\citep{tali2024flowbenchlargescalebenchmark,choudhary2025pregeneratingmultidifficultypdedata}, ours crucially is generated via the GPU-native PyFR code~\citep{PyFR}, enabling a direct and fair comparison to neural network training and inference on the same device type.\looseness-1

\vspace{-3mm}
\section{Breakeven Complexity}
\label{sec:breakeven}
\vspace{-2mm}

We are interested in solving $\theta$-parameterized PDEs of form \looseness-1
$\partial_tu(t,x)=\mathcal L_\theta[u](t,x)$
with initial condition 
$u(0,x)=u_\theta(x)$
and boundary terms
$\B_\theta[u|\partial\X](t,x)=0$
over a domain $\X\ni x$,
with design variable $\theta$ parameterizing the differential operator $\mathcal L_\theta$, the initial conditions $u_\theta$, and the boundary conditions $\B_\theta$.  
The goal of a PDE solver is to output an approximate solution $\hat u_\theta$ over $\X$, with the approximation quality assessed either by comparing to a high-fidelity reference or via a PDE residual.
Traditionally, transient PDEs have been solved by what we will refer to as ``classical'' solvers, which typically discretize in space and time and solve a sequence of numerical problems on the resulting grid.
We will use such solvers to generate our data and high-fidelity reference solutions for evaluations.\looseness-1

\vspace{-2mm}
\subsection{Neural PDE Solvers}
\vspace{-1mm}
A neural PDE solver is one where the values of the function $\hat u_\theta(t,x)$ at specific space--time points are computed using a neural network $f_w$ with weights $w\in\mathbb{R}^n$, i.e.\ $\hat u_\theta(t,x)=f_w(\theta,t,x)$.
They are often trained by sampling a set of $N_{\mathrm{data}}$ points $\theta_i,i=1,\dots,N_{\mathrm{data}}$, from some distribution $\mathcal D$ over $\Theta$, classically computing corresponding ``ground truth'' solutions $u_i$ for each of them, and training $f_w$ to approximate these solutions.
We will evaluate the performance of learned (and other non-reference) solvers via the {\bf normalized RMSE~(nRMSE)} $\|f_w(\theta,\cdot,\cdot)-u_\theta\|_2/\|u_\theta\|_2$ relative to a ground truth reference solution, averaged over output channels and over a held-out test set of points $\theta\sim\mathcal D$.
This serves as an imperfect indicator for the performance of the method on solving downstream tasks, e.g. inverse problems over the domain $\Theta$.\looseness-1

\vspace{-2mm}
\subsection{Motivation: Where are neural solvers used?}
\vspace{-1mm}
The scientific workloads where neural PDE solvers have the most promise involve \emph{repeated} forward solves of related PDEs, often to optimize some task metric.
For example, solving a design optimization or inverse problem task with an objective like
$
\label{eq:invprob}
    \theta^\star \in \arg\max_{\theta \in \Theta}\; J(\theta; u_\theta).
$
typically requires many forward solves at different values of $\theta$, which is where acceleration via learned solvers may be useful. 
To quantify this, suppose $N$ forward PDE solves with accuracy $\varepsilon$ suffices to optimize this objective to an acceptable value.
Then the question we seek to answer next is whether it is better to use a classical or neural solver for this purpose.\looseness-1

\vspace{-2mm}
\subsection{Breakeven Complexity}
\vspace{-1mm}
A key feature of this setting is the up-front cost of a neural PDE solver $f_w$, which is the sum of two quantities:
(i)~the cost of generating the data and (ii)~the cost of training the model on that data.
Since data generation cost can be reasonably assumed to scale linearly with the number $N_{\mathrm{data}}$ of training points, we write the data generation cost as $C_{\mathrm{gen}} N_{\mathrm{data}}$, where $C_{\mathrm{gen}}$ is the average cost of generating one training trajectory via a classical solver.
We then define the total up-front training cost as
$
\label{eq:train_cost}
    B := C_{\mathrm{gen}} N_{\mathrm{data}} + C_{\mathrm{train}},
$
where $C_{\mathrm{train}}$ is the total cost of all training steps. $B$ can be interpreted as the budget we spend on producing a neural solver before it can be deployed.
Let $\varepsilon_B$ denote the (worst-case or average) error achieved by the resulting model.\looseness-1

Now, if our downstream task needs $N$ trajectories with error $\varepsilon_B$ to solve, then the total cost spent on the task will be
$
\label{eq:total_cost}
    B_{\mathrm{total}} := B + C_\mathrm{inf}N,
$
i.e. the sum of the up-front training cost $B$ and total inference cost $C_\mathrm{inf}N$, where $C_\mathrm{inf}$ is  cost per inference using $f_w$.\looseness-1

\begin{figure}[!t]
  \centering
  \begin{minipage}{0.48\textwidth}
    \centering
    \includegraphics[width=\linewidth]{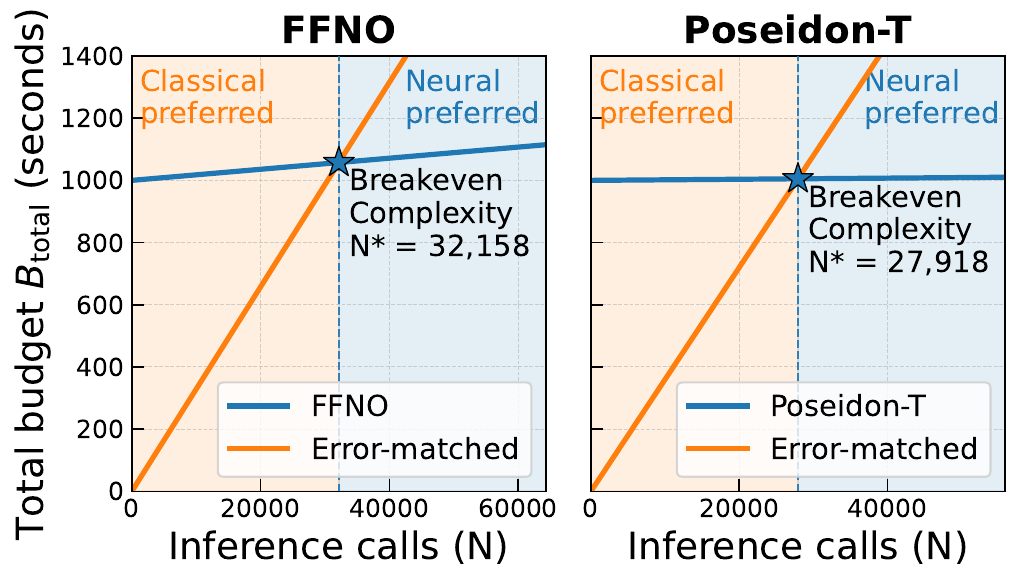}
    \caption{Illustration of the breakeven complexity definition, using FFNO and Poseidon-T on Navier-Stokes as examples.
    For a fixed training budget $B=1000$, a learned solver's total costs for simulating $N$ trajectories is $B_\textrm{total} = B + C_{\textrm{inf}}N$, while the cheapest classical configuration whose error matches the learned accuracy $\varepsilon_B$ has cost $C_{\delta_B}N$. The intersection of these two lines, if ever, will occur at $N^\star(B)=\frac{B}{C_{\delta_B}-C_{\mathrm{inf}}}$ (Eq. \ref{eq:breakeven}), which we define as the breakeven complexity.
    For any problem that requires fewer inference calls than this, we prefer to use the classical solver.}
    \label{fig:breakcomplexityintersect}
  \end{minipage}
  \hfill %
  \begin{minipage}{0.48\textwidth}
    \centering
    \includegraphics[width=\linewidth]{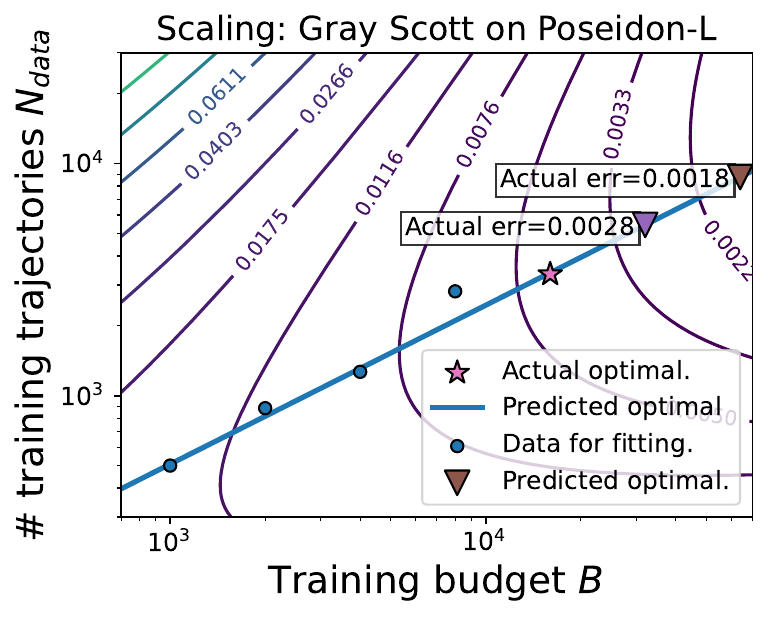}
    \caption{Example of a scaling landscape for the budget-optimal error, using Poseidon-L trained on Gray-Scott and varying the allocation between training trajectories (data-generation) and optimization steps. We fit a model to the small-budget optima and extrapolate to larger budgets to estimate the best-achievable error at each $B$. A detailed fitting procedure is explained in Appendix~\ref{Adx:scaling}.\looseness-1}
    \label{fig:scaling}
  \end{minipage}
\end{figure}

To fairly compare a neural PDE solver against a classical solver~\cite{weak_baseline}, we compare $B_\mathrm{total}$ to the cost that a similarly accurate classical solver would incur in order to compute $N$ trajectories.
Specifically, let $\delta_B$ denote the fidelity---e.g. the resolution---of an {\bf error-matched} classical solver, i.e. one that achieves the same (worst-case or average) error $\varepsilon_B$ as the model, and let $C_{\delta_B}$ denote its per-trajectory cost.
Then, the neural PDE solver is cheaper at the attained error if and only if, for $N$ required trajectories,
$
\label{eq:break1}
    B + C_{\mathrm{inf}}N < C_{\delta_B}N.
$
Rearranging the equation yields an amortization threshold, which we define as the {\bf breakeven complexity}:\looseness-1
\begin{equation}
\label{eq:breakeven}
    N^\star := \frac{B}{\max \{ C_{\delta_B} - C_{\mathrm{inf}}, 0\}}.\vspace{-1mm}
\end{equation}
As illustrated in Fig.~\ref{fig:breakcomplexityintersect}, breakeven complexity is the point when a neural PDE solver and its error-matched classical solver require the same cost of solving the downstream task.
It thus isolates the amortization regime, being small when (i)~the up-front cost $B$ is modest, (ii)~inference is substantially cheaper than that of the error-matched classical simulation ($C_{\delta_B}\gg C_{\mathrm{inf}}$), or both. 
Conversely, breakeven complexity is infinite when classical low-fidelity simulation is already cheaper than inference at matched accuracy.\footnote{One could choose to also set $N^\star=\infty$ if the nRMSE is greater than one, since that is attained by an all-zero predictor.
While we choose to always error-match with a downscaled solver, this observation highlights that $N^\star$ should be assessed together with the error metric it is based on, since that ultimately determines whether a model is suitable for the task.}
One limitation here is the assumption that simulation costs are roughly the same for all parameters $\theta$, which could fail in certain settings where $\theta$ influences the difficulty of the problem;
however, taking the average cost over a distribution is a reasonable and meaningful approximation.\looseness-1

\vspace{-3mm}
\section{Evaluation Methodology}
\label{sec:eval}
\vspace{-2mm}
In this section, we describe the experimental evaluation protocol we use to compute breakeven complexity. 
The pipeline involves (i)~determining the cost measure, (ii)~estimating the best achievable neural solver error for a fixed training budget, (iii)~matching the latter's accuracy to a classical baseline's, and (iv)~computing the breakeven complexity via the quotient of the training budget and the difference between neural and traditional generation cost.\looseness-1

\vspace{-2mm}
\subsection{Measuring compute cost}
\vspace{-1mm}
We use the wallclock time~(seconds) to measure cost.
Although FLOPs are commonly used for scaling laws, they are less useful for comparing neural and classical PDE solvers because (i)~throughput (FLOPs/s) depends strongly on kernel structure/hardware utilization and (ii)~classical solver costs are dominated by stencil/flux evaluations, memory effects, and communications that are not well-represented via FLOPs. 
Wallclock therefore provides a direct end-to-end cost currency shared by both learned and classical methods, one which is also used in related work~\cite{ashton2025fluid}. We exclude training initialization overheads (e.g. compilation and warmup) as they are not fundamental to the underlying modeling or physics problems.
Lastly, in this paper we run both neural and traditional solvers on the same hardware and ensure high GPU-utilization in both cases;
in future settings where classical simulation must be done on CPU, a monetary or power usage comparison could be used.

\vspace{-2mm}
\subsection{Error-optimal data generation}
\vspace{-1mm}
A learned solver incurs up-front cost $B = C_{\textrm{gen}} N_{\textrm{data}} + C_{\textrm{train}}$ from data generation and training,
where $C_{\textrm{gen}}$ is per-trajectory generation time, $N_{\textrm{data}}$ is the number of generated trajectories, and $C_{\textrm{train}}$ is total training step time.
For a given budget $B$, neural solvers admit many feasible allocations (more data vs.\ more optimization, different model sizes/hyperparameters) that can yield substantially different errors and inference costs. Since breakeven complexity compares methods at a \emph{fixed up-front budget}, using a non-optimal configuration would conflate the metric with arbitrary training choices. We therefore estimate the \emph{best achievable} error under each budget and use this budget-optimal model as the representative learned solver at budget $B$.\looseness-1

Empirically, this can be done by sweeping feasible model configurations (i.e. data generation and optimization allocation) for some fixed budget. However, since sweeps at large budgets can be costly, we instead take the following scaling laws-inspired approach. 
First, we run hyperparameter sweeps on a set of small budgets $B$ while also trying different generation vs. optimization allocations (varying $N_\textrm{data}$ vs. $C_\textrm{train}$ while keeping $C_\textrm{gen}N_\textrm{data}+C_\textrm{train}\le B$).
We set $\varepsilon_B$ to be the best (worst-case or average) error achieved at small budget $B$ and define $N_\textrm{data}(B)$ to be the associated optimal number of training trajectories.
Then we fit a scaling model to the labeled pairs $(B,N_\textrm{data}(B))$ in order to predict the error-optimal amount of data to generate at larger budgets $B$~\cite{hoffmann2022trainingcomputeoptimallargelanguage}. An example of the fitted scaling landscape graph is show in Fig.~\ref{fig:scaling}.

\vspace{-2mm}
\subsection{Error-matched classical solvers}\label{sec:error-matching}
\vspace{-1mm}
While high-fidelity classical solvers are typically much more expensive than neural PDE solvers, the latter often have much higher error;
thus a fair comparison is to compare to low-fidelity classical solver.
Thus, for each training budget $B$, we find the cheapest configuration of the classical solver whose error matches that of the learned solver's accuracy $\varepsilon_B$ and use that as our baseline.
This error-matching is essential: classical solvers can often be run at reduced fidelity (and thus much lower cost), so comparing surrogate inference cost to an unnecessarily high-fidelity baseline can overstate the usefulness of neural PDE solvers~\citep{weak_baseline}.\looseness-1

More formally, let $\{\mathsf S_\delta\}$ be a family of classical solver configurations indexed by fidelity $\delta$, with per-trajectory runtime $C_\delta$. 
Let $\varepsilon(\delta)$ denote its error, computed in the same way as the neural PDE solver error by comparing to a high-fidelity reference on a set of test trajectories.
Then we define the {\bf error-matched solver} to be the cheapest solver that is at least $\varepsilon_B$-accurate: \looseness-1
\begin{align}
\label{eq:match}
\delta_B &\in \arg\min_{\delta}\Big\{ C_\delta:\ \varepsilon(\delta)\le \varepsilon_B\Big\}
\end{align} \looseness-1
In practice, we can construct the family $\{\mathsf S_\delta\}$ of low-fidelity solvers by coarsening the resolution and timesteps.
Starting from a high-fidelity solver, we decrease the resolution by e.g. setting grid spacing $\Delta x \to r \Delta x$ for $r\in\{2,4,\dots\}$ in the case of regular grids or analogously reducing the number of nodes in the case of unstructured meshes. 
For each spatial coarsening level, we increase the time step $\Delta t$ to reduce runtime, while maintaining stability by enforcing a CFL~(Courant–Friedrichs–Lewy) condition constraint.
Concretely, we choose $\Delta t$ so that the nondimensional Courant number remains approximately constant across fidelities, i.e.
$
\mathrm{CFL} \;\propto\; \frac{u_{\max}\,\Delta t}{\Delta x} \approx \mathrm{const},
\label{eq:cfl_const}
$
where $u_{\max}$ is a characteristic velocity scale for the benchmark equation.
When the solver has additional stability constraints (e.g., diffusion-like terms), we additionally cap $\Delta t$ by the corresponding bound.
Note that, for some numerical methods (e.g. ETDRK time-steppers \cite{fu2024higherorderenergydecreasingexponentialtime, xu2025stabilitytimestepconstraintsexponential}), the stability and CFL condition does not depend significantly on $\Delta t$. In this case, we would fix $\Delta t$ and only coarsen the grid resolution.
This procedure yields a principled grid-timestep schedule that produces a monotone cost-accuracy tradeoff and avoids unstable configurations.

\vspace{-2mm}
\subsection{Computing breakeven complexity}
\label{sec:compute_breakeven}
\vspace{-1mm}
After determining the error-matched solver by defining a family of low-fidelity solvers and using Eq.~\ref{eq:match}, we are finally able to compute the breakeven complexity~\eqref{eq:breakeven} $N^\star(B)=B/\max\{C_{\delta_B}-C_\textrm{inf},0\}$, 
with $N^\star(B)=\infty$ when the denominator is 0, i.e. when inference cost is no cheaper than classical simulation.
Note that, given a static set of test trajectories, we can define the error $\varepsilon_B$ of the neural PDE-solver and the error $\varepsilon(\delta_B)$ of its matched classical solver in (at least) two ways:
via the {\bf average nRMSE} over the test set or the {\bf worst nRMSE} of any trajectory in the test set.
We use these to define the {\bf average-case breakeven complexity} and the {\bf worst-case breakeven complexity}, respectively.
Average-case breakeven complexity indicates the workload size required for a learned solver to amortize training cost to be more useful than classical simulators when targeting typical accuracy, while worst-case breakeven complexity indicates the workload size required to amortize training cost while meeting robust worst-trajectory accuracy guarantees.
Note that since we consider the worst-case for the both the neural solver and the error-matched classical solver, the worst-case breakeven complexity is not necessarily worse than the average-case.\looseness-1

\vspace{-3mm}
\section{Benchmarks and datasets}\label{sec:benchmarks}
\vspace{-2mm}

\begin{figure*}[!t]
\centering
\includegraphics[width=\linewidth]{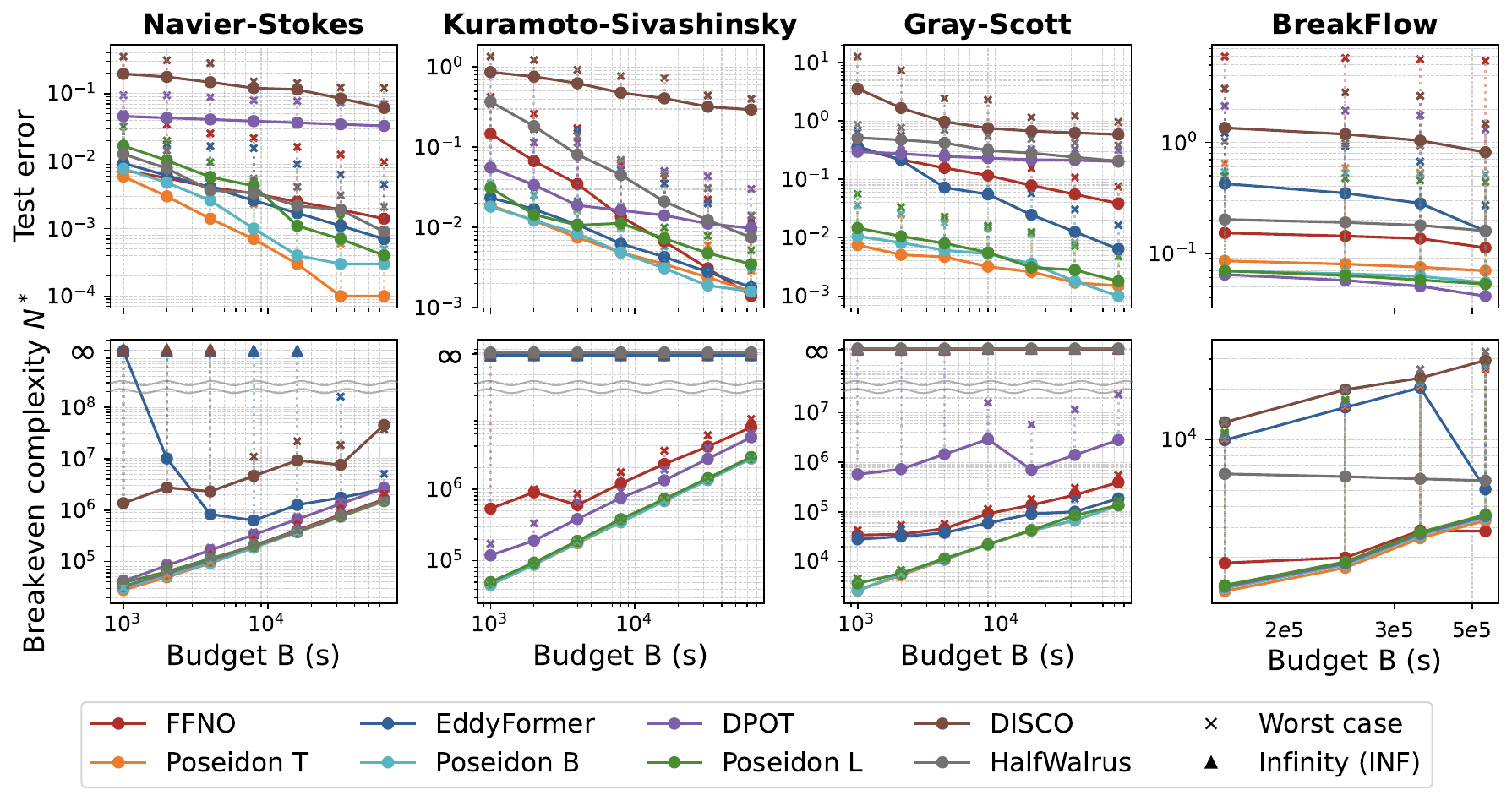}
\caption{Plots of the test nRMSE~(top) and breakeven complexity~(bottom), both as functions of the training budget $B$.
``$\times$'' and ``$\blacktriangle$'' represent worst-case~(error or complexity) and infinite~(complexity), respectively.
While error decreases monotonically almost everywhere, breakeven complexity is nonlinear because both its numerator and denominator depend on the budget $B$.
\looseness-1}
\label{fig:result}
\end{figure*}

 We evaluate breakeven complexity on six PDE families:
five extant settings from APEBench~\cite{koehler2024apebench} and a new benchmark, BreakFlow, which uses PyFR~\citep{PyFR} to simulate flows in complex geometries.
The suite covers smooth, dissipative transient dynamics, with simulations run until the evolving quantities converge to a steady state, enter limit cycles, or reach statistically stationary behavior.
Details of the benchmarks and datasets are presented in Appendix~\ref{Adx:data}. \looseness-1

The {\bf periodic benchmarks} we consider involve three canonical equations: Navier-Stokes~(N-S), Kuramoto-Sivashinsky~(K-S), and Gray-Scott~(G-S).
They are chosen due to their nonlinearity and  distinct physical characteristics (advective turbulence, chaotic instability-driven dynamics, and reaction–diffusion pattern formation, respectively). 
While our primary evaluations focus on 2D domains, we also evaluate the K-S equation in 1D and 3D to analyze difficulty scaling across spatial dimensions.
The data trajectories are simulated via the Exponax solver from APEBench~\cite{koehler2024apebench}, a GPU-based pseudo-spectral solver with ETDRK time-stepping. Its Fourier spectral discretization is well-suited to smooth periodic dynamics and provides an efficient, well-controlled classical baseline for breakeven analysis.\looseness-1

The {\bf BreakFlow benchmark} is a new setting we introduce consisting of flows past multiple objects simulated in PyFR~\citep{PyFR}, a flux-reconstruction code that uses artificial viscosity to enable highly parallel GPU operations for incompressible flows in complex geometries;
this makes it particularly well-suited for breakeven analysis.
The specific PDE instances in BreakFlow involve rightward flows past rectangles, with sizes, positions, and rotations sampled from a distribution ensuring sufficient separation between obstacles and symmetry about the horizontal axis, as detailed in Appendix~\ref{Adx:data}. 
We generate simulations over three different ranges of Reynolds numbers, $(10, 40]$, $(40, 90]$ and $(90,160]$.\looseness-1

\vspace{-3mm}
\section{Experimental Results}
\vspace{-2mm}

We use the evaluation pipeline in Sec.~\ref{sec:eval} to compute breakeven complexity across the four PDE setups in Sec.~\ref{sec:benchmarks} for multiple model classes.
Specifically, we evaluate two categories of neural PDE solvers: \emph{supervised} models trained from scratch on each task and \emph{foundation-model}~(FM) approaches adapted to each task via fine-tuning.
For the former we consider four representative architectures: 
(i)~\emph{FFNO}~\cite{ffno}, an efficient and tuned variant of the Fourier Neural Operator~\citep{fno}, (ii)~\emph{EddyFormer}, a Transformer-based model~\cite{eddyformer}, 
(iii)~\emph{DISCO}, which uses a hypernetwork to dynamically generate an autoregressive operator~\cite{disco}, and
(iv)~\emph{DPOT}, which uses a Fourier-based attention mechanism~\cite{dpot}.
For the latter, we consider two representative FM architectures: (i)~the \emph{Poseidon} family (Tiny, Base, and Large) pre-trained on diverse fluid data~\cite{poseidon}, and (ii)~\emph{HalfWalrus}, a Transformer-based cross-domain FM pre-trained on a broad set of continuum dynamics from many physical scenarios~\cite{walrus}.
When computing breakeven complexity, we exclude pretraining cost, treating it as a shared, reusable expense amortized across many downstream tasks and thus not attributable to any single deployment.\looseness-1

A complete report of our breakeven complexities at different budgets can be found in Appendix~\ref{Adx:table}, with model training details provided in Appendix~\ref{Adx:train}.
Note that we compute both \emph{average-case} and \emph{worst-case} variants (Sec.~\ref{sec:compute_breakeven}), where worst-case is taken over test trajectories.
In this section, we organize our experimental analysis around the following three claims:\looseness-1
\vspace{-3mm}
\begin{enumerate}[noitemsep, leftmargin=*]
    \item By forcing the consideration of compute cost, breakeven complexity reveals relative model strengths and weaknesses overlooked in previous accuracy-driven evaluations.
    \item We can compare worst-case and average-case variants of breakeven complexity to gain an understanding of the robustness of a learned model relative to a low-fidelity classical solver.\looseness-1
    \item Breakeven complexity evaluations reveal that neural solvers need to be called hundreds of thousands of times to be effective in the basic toy PDEs that are typically studied in scientific ML, but on the other hand neural solvers become {\em more} cost-effective on {\em harder} PDE problems.
\end{enumerate}

\begin{figure}[!t]
  \centering
  \begin{minipage}{0.58\textwidth}
    \centering
    \includegraphics[width=\linewidth]{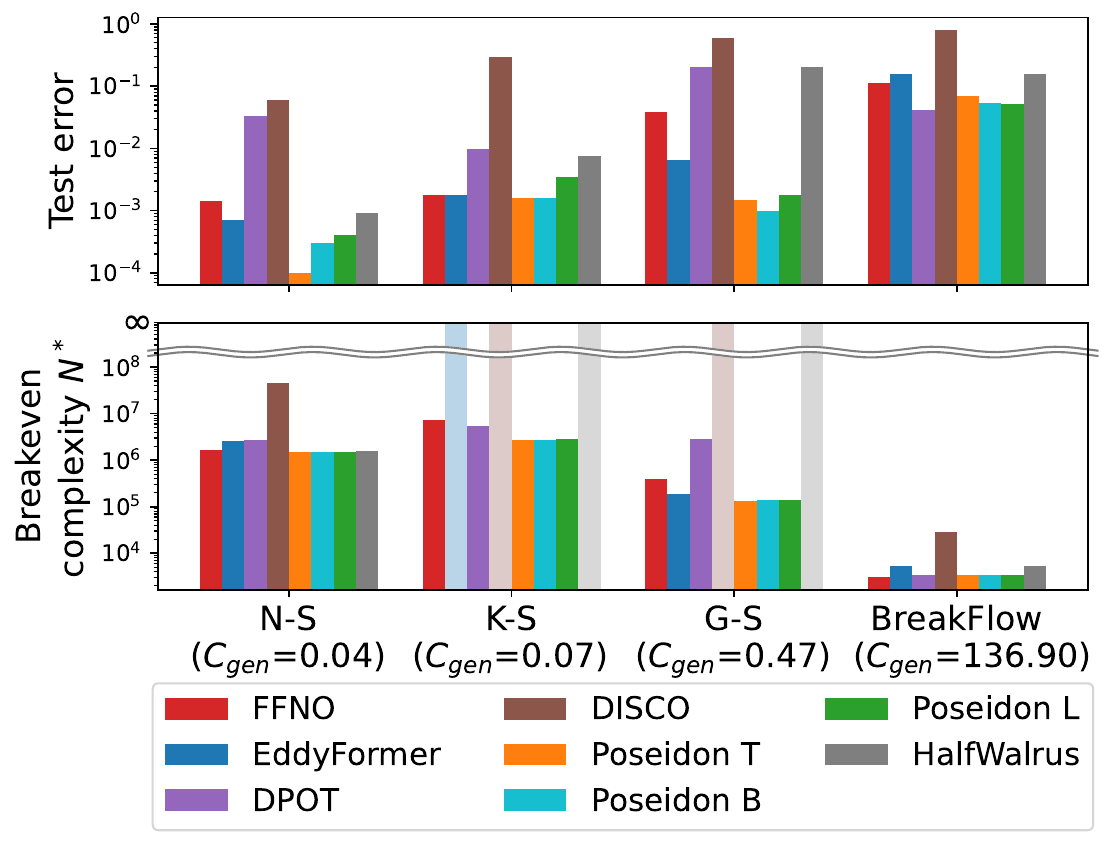}
    \caption{
    nRMSE~(top) and breakeven complexity~(bottom) for different models across the four tasks at the largest budget.
    While nRMSE suggests neural surrogates will struggle on (classically) demanding tasks given the same (training) budget, our compute-aware metric suggests the opposite, being lower for PDEs with longer simulation time~($C_\textrm{gen}$).\looseness-1
    }
    \label{fig:be_bar}
  \end{minipage}
  \hfill %
  \begin{minipage}{0.38\textwidth}
    \centering
    \includegraphics[width=\linewidth]{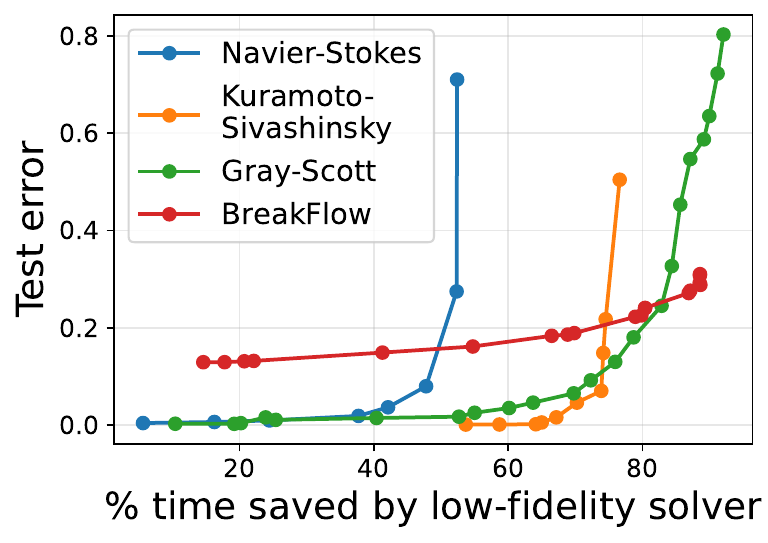}
    \vspace{1mm}
    \caption{Comparison of how decreasing resolution increases simulation speed across our four PDE settings;
    the time saved is relative to the high-fidelity solver, with the sequence of low-fidelity solvers determined as in Section~\ref{sec:error-matching}.
    Unlike the APEBench simulations, BreakFlow is extremely sensitive to changes in resolution, with error increasing rapidly after a slight decrease in solve cost.\looseness-1
    }
    \label{fig:lowfidcompare}
  \end{minipage}
\end{figure}

\vspace{-3mm}
\subsection{Claim 1: The value of cost-awareness}
\vspace{-1mm}
In Fig.~\ref{fig:result} we plot the budget-optimal test error $\varepsilon_B$ and the corresponding breakeven complexity $N^\star(B)$ as functions of the training budget $B$, doing so across all models and settings under consideration.
While the error curves are almost uniformly decreasing across the models and settings, the breakeven complexity can be non-monotone;
notably, EddyFormer struggles on low budgets for both fluid settings (Navier-Stokes and BreakFlow) but becomes much stronger at higher budgets, suggesting that to be effective it must be trained sufficiently long and on sufficiently many samples.
On the other hand, EddyFormer has $N^\star=\infty$ on Kuramoto--Sivashinsky because its inference cost is greater than the simulation cost of an error-matched classical solver, which permits substantial coarsening before its error rises sharply (Fig.~\ref{fig:lowfidcompare}).
The same holds for two of the APEBench settings for HalfWalrus, owing to its high inference cost (Appendix~\ref{Adx:train}).
A surprising finding is that while the Poseidon models and DPOT dominate BreakFlow according to test error, their breakeven complexity is not much better than that of HalfWalrus and FFNO, with the latter being the most cost-effective at the highest budget we evaluate despite being a supervised model.\looseness-1

Beyond this, breakeven complexity's forced consideration of cost tradeoffs reveals performance shifts that standard metrics might overlook.
Most notably, when training cost is equalized we find that Poseidon-L is the worst FM in its family on all tasks except BreakFlow, according to both breakeven complexity (albeit barely) and test nRMSE (more significantly).
In contrast, it was reported to be far better than Poseidon-T and Poseidon-B in aggregate on similar tasks~\citep[Table~9]{poseidon}.
This demonstrates the usefulness of our new measure in forcing an accounting of compute costs when doing evaluation.\looseness-1

\begin{figure*}[!t]
\centering
\includegraphics[width=\linewidth]{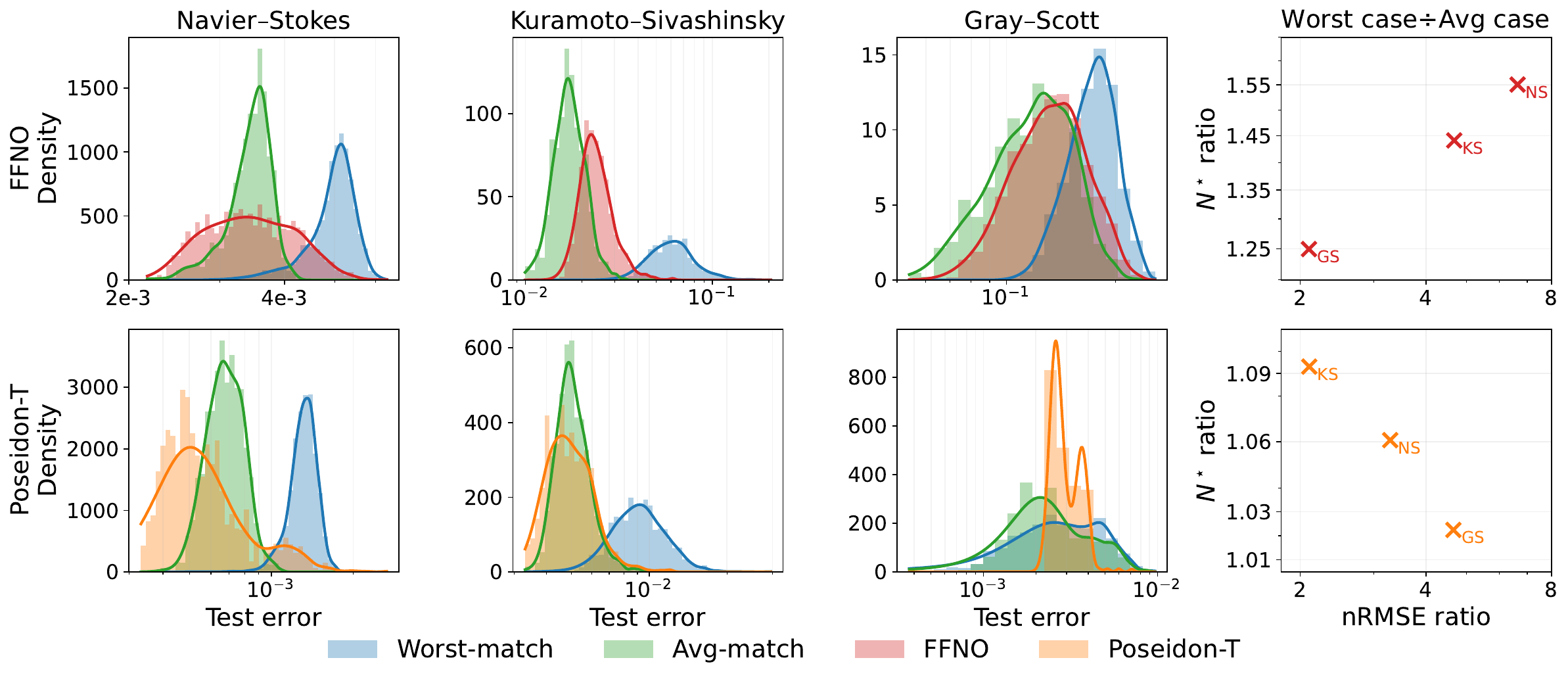}
\caption{
Comparison of the error distributions of FFNO and Poseidon-T (training budget 8K) along with the distributions of the classical solvers that match their worst-case and average-case performance.
On the right we compare the worst-case to average-case ratios for $N^\star$ and nRMSE.
For both models, their error distribution on Gray-Scott is much more robust, i.e. more concentrated, than on the other PDEs, making the distribution of the worst-case match solver fairly close to that of the average-case;
this in turn means the worse solver is not much faster and so the worst-case breakeven complexity does not increase so much over its average-case.
This is evidenced in the plots on the right, where the ratio of worst-to-average-case complexity is smaller for Gray-Scott.
Notably, the ratio of the worst-case to the average-case nRMSE does not correlate to robustness in this way.\looseness-1}
\label{fig:distribution}
\end{figure*}

\vspace{-2mm}
\subsection{Claim 2: Measuring robustness}
\vspace{-2mm}
As discussed in Section~\ref{sec:breakeven}, we can measure breakeven complexity by matching learned and traditional solvers via their {\em average} test error or their {\em worst-case} test error.
As shown in Figure~\ref{fig:distribution}, it is instructive to look at their ratio:
both FFNO and Poseidon-T have relatively close average and worst-case complexities on Gray-Scott, but on the other PDEs the worst-case tends to be much higher.
Looking at the distribution of errors across the test trajectories of each task in Fig.~\ref{fig:distribution}, we see that Gray-Scott is also the setting where the surrogates' errors are more tightly concentrated than those of their matched solvers. 
This suggests a smaller gap between worst-case and average-case breakeven complexity reflects a more robust learned solver, i.e. one that is less likely to have instances where it performs much worse than the average instance.
Mathematically, this will be caused by the worst-case error-matched solver having an error distribution not too far from the average-case, thus also not being much cheaper and therefore not significantly increasing the complexity.
Since optimization problems naturally require exploring unseen parts of the parameter space, the fact that breakeven complexity, unlike nRMSE, can be indicative of robustness is practically meaningful.\looseness-1

\begin{figure*}[!t]
\centering
\includegraphics[width=\linewidth]{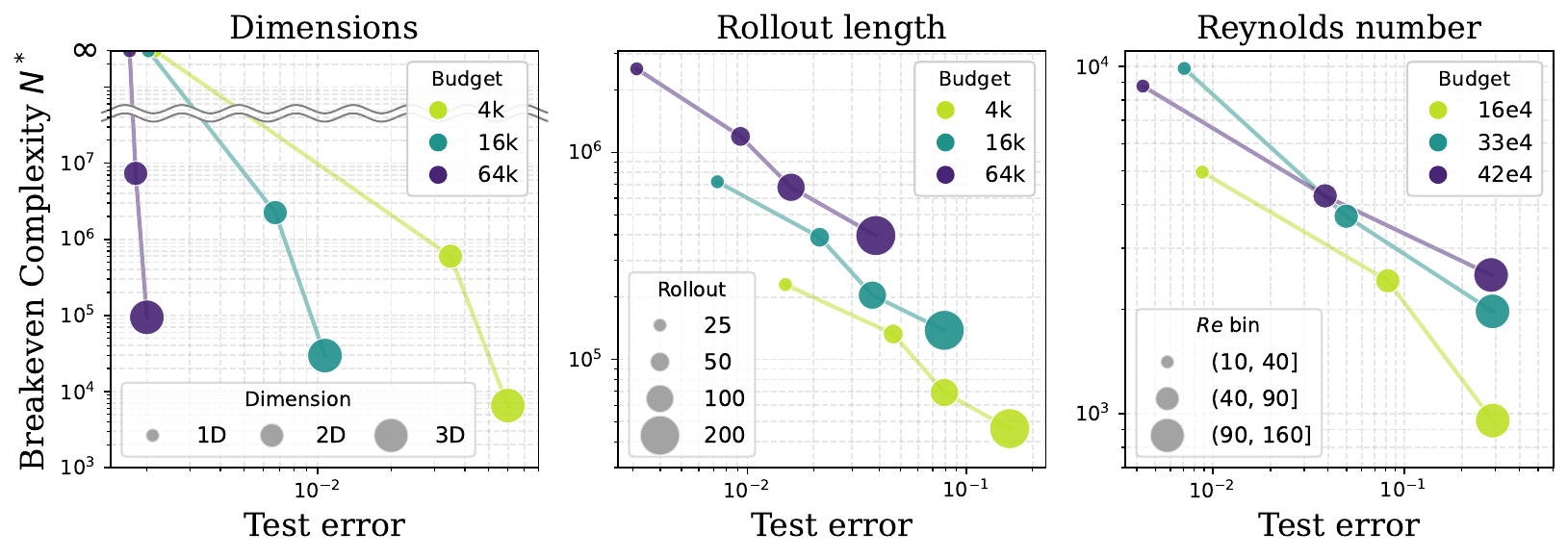}
\includegraphics[width=\linewidth]{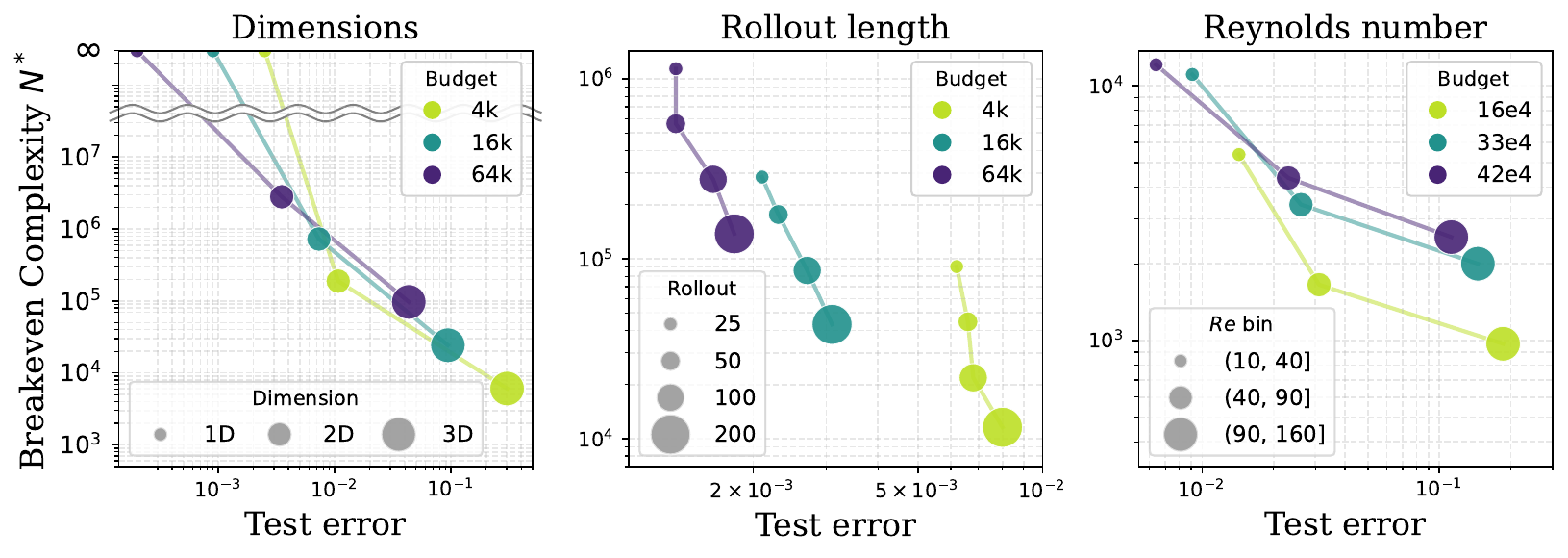}
\caption{Plot showing the inverse relationship between test error and breakeven complexity of FFNO (top) and Poseidon-L (bottom) as problem difficulty increases on Kuramoto-Sivashinsky~(left, varying spatial dimension), Gray-Scott~(middle, varying rollout length), and BreakFlow~(right, varying Reynolds number).
Across all three problem settings, increasing problem difficulty sharply reduces breakeven complexity at the same budget, indicating a growing computational advantage for neural surrogates over classical solvers.\looseness-1}
\label{fig:difficulty}
\end{figure*}

\vspace{-2mm}
\subsection{Claim 3: Optimism in the face of harder PDE problems}
\vspace{-2mm}

At first glance, our evaluations in toy settings suggest that hundreds of thousands of calls are necessary to reach cost-effectiveness against error-matched traditional solvers (Fig.~\ref{fig:be_bar}). 
However, the same plot demonstrates that breakeven complexity scales inversely with the computational cost of classically solving the PDEs, a basic notion of problem difficulty.
In particular, neural surrogates typically require only a few thousand inference calls to break even on BreakFlow, an order of magnitude fewer than on Gray-Scott (for most models) and two orders of magnitude fewer than on Navier-Stokes and Kuramoto-Sivashinsky.
On the other hand, simulating BreakFlow is two orders of magnitude more expensive than G-S and three orders more than N-S and K-S.
This trend suggests that the utility of neural PDE solvers may lie largely in high-fidelity regimes.\looseness-1

However, scaling with simulation cost does not on its own explain this trend, as there could be other reasons why learned solvers may have such a dramatically smaller breakeven complexity on BreakFlow.
For example, Figure~\ref{fig:lowfidcompare} shows that on the APEBench settings we can easily decrease the solver cost without significantly increasing error, while on BreakFlow the solver is more sensitive, with error rising dramatically even when the low-fidelity solver is only $10\%$ faster than the high-fidelity one.
In such cases, the simulation cost of the error-matched solver will remain large and thus will dominate the denominator of Eq.~\ref{eq:breakeven}, making breakeven complexity small.\looseness-1

We thus conduct further analysis along three other, better-defined difficulty axes: spatial dimensionality (on K--S), temporal rollout length (on G--S), and flow complexity in terms of Reynolds number $Re$ (on BreakFlow). As shown in Figure~\ref{fig:difficulty}, for both FFNO and Poseidon-L models, all three settings exhibit the same pattern observed above: as problem difficulty increases while the budget is kept fixed, the breakeven complexity {\em decreases} while the test error {\em increases}. In more detail, moving from 1D to 3D on Kuramoto--Sivashinsky substantially increases classical solver cost, lowering the breakeven threshold despite the increase in surrogate error. For Gray--Scott, longer rollout horizons make coarse classical solvers less viable because temporal errors accumulate, requiring finer resolutions and again reducing breakeven complexity. Lastly, BreakFlow simulations with well-formed vortex streets ($Re\in(90,160]$) require finer grids and more restrictive time steps than steady flows ($Re\in(10,40]$), yielding the same trend along the flow-complexity axis~\cite{williamson1989oblique,roshko1954development}.

\begin{figure*}[!t]
\centering
\includegraphics[width=\linewidth]{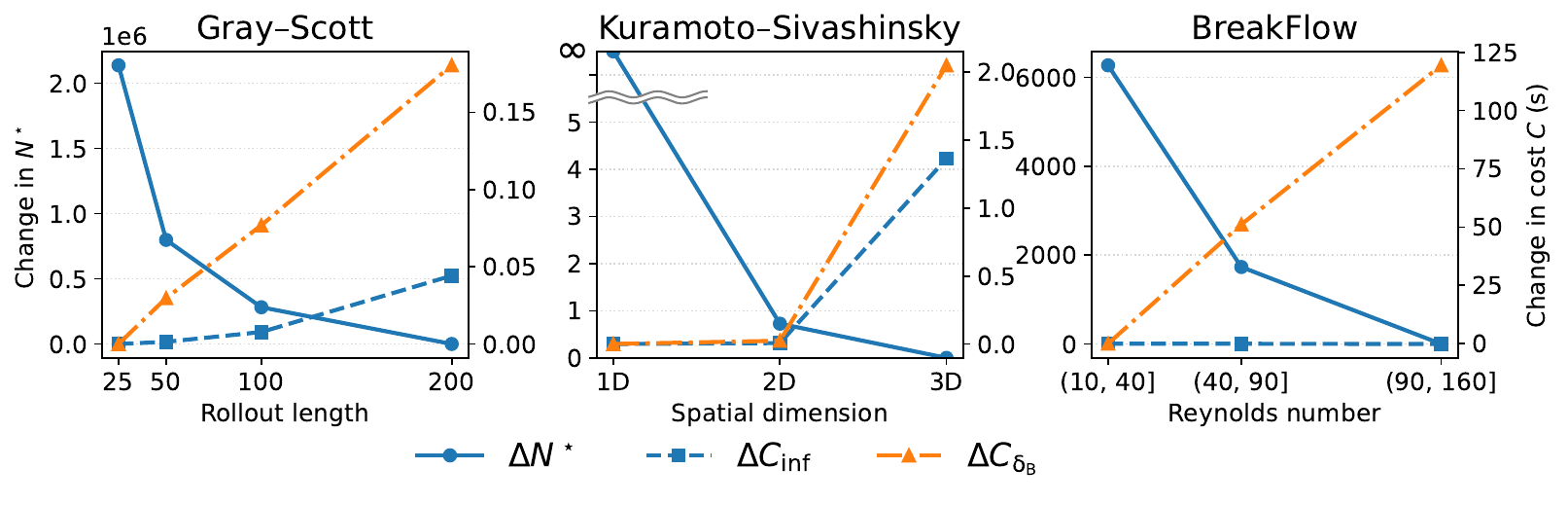}
\caption{Decomposition of breakeven complexity across difficulty levels for FFNO at the largest evaluated budget (64K seconds for the periodic settings and 420K seconds for BreakFlow). We report changes in $N^\star$ (left axis), neural inference cost $C_{\mathrm{inf}}$ (right axis, blue), and error-matched classical solver cost $C_{\delta_B}$ (right axis, orange) across three difficulty axes (rollout length, spatial dimensionality, and physical complexity measured by Reynolds number). We see that, across all three settings, the classical solver cost increases substantially faster than the neural inference cost.\looseness-1}
\label{fig:infer_analysis}
\end{figure*}

To better understand this trend, we further decompose the breakeven complexity at the largest evaluated budget over the FFNO difficulty ablations. Since $B$ is fixed within each difficulty sweep, a decrease in $N^\star$ is driven by an increase in the per-trajectory saving $C_{\delta_B}-C_{\mathrm{inf}}$. Fig. \ref{fig:infer_analysis} shows that across all three difficulty axes, $C_{\delta_B}$ grows substantially more rapidly with difficulty than $C_{\mathrm{inf}}$. 
Thus, on the harder problems we tested, the learned solvers save more time per trajectory relative to the error-matched classical solver and break even in fewer trajectories, despite their higher test errors.

These results suggest that neural surrogates can be most computationally advantageous in regimes where classical solver costs grow rapidly, i.e., on more challenging, large-scale problems. As these are precisely the regimes of interest in practical scientific computing, breakeven complexity provides a more optimistic view of the potential of ML for scientific computing than that suggested by \citet{weak_baseline}, despite their initial introduction of the metric.
\looseness-1

\vspace{-3mm}
\section{Conclusion} 
\vspace{-2mm}

We formalize \emph{breakeven complexity}, a framework for evaluating neural PDE solvers that complements standard error-based evaluation by quantifying the threshold at which a learned PDE solver becomes cost-effective relative to an error-matched classical baseline; 
it thus integrates end-to-end costs (data generation, training, inference) while accounting for low-fidelity solvers.
Across three 2D periodic PDEs and the BreakFlow benchmark, the metric reveals hidden weaknesses in past evaluations, naturally quantifies robustness, and suggests that the greatest potential of neural PDE solvers will be realized on more challenging benchmarks.
As a result, we believe the metric should be considered by any rigorous evaluation of neural PDE methods and should also inform future benchmark construction. The metric can temper overoptimistic speedup claims and thus may help researchers with limited compute choose affordable, reliable solvers. \looseness-1

{\bf Limitations} of breakeven complexity include its dependence on the hardware, the quality of the classical baseline implementation, and the modeling of inference and training costs; this makes it more challenging to work out than simple error statistics. However, such analysis is exactly what enables a practitioner to assess its value for their use case.
Our evaluation is also limited in coverage, with no (e.g.) out-of-distribution settings, shock-dominated dynamics, or stationary problems.

{\bf Future work} may thus extend the framework or its evaluation to such regimes, or incorporate heterogeneous hardware, adaptive classical solvers, or uncertainty in workload forecasts. 
Other directions include combining the metric with concrete inverse problems and defining it for the case of hybrid solvers that use learning to correct low-fidelity traditional solvers~\citep{kochkov2021machine,eddyformer}.\looseness-1

\bibliography{reference}
\bibliographystyle{chicago}

\appendix
\onecolumn

\section{Dataset specifications}
\label{Adx:data}

All data generation uses FP32 on NVIDIA L40S GPUs with no concurrent jobs on the GPU.

\paragraph{2D incompressible Navier-Stokes}
We consider this equation in vorticity form.
Letting the velocity field be $\mathbf u(t,x)=(u_1,u_2)$ with $\nabla\cdot \mathbf u = 0$, define the scalar vorticity
$\omega(t,x) := \partial_{x_1} u_2 - \partial_{x_2} u_1$.
The vorticity equation is
\begin{align}
\label{eq:ns}
    \partial_t \omega + \mathbf u \cdot \nabla \omega = \nu \Delta \omega + f,
\end{align}
where $\nu>0$ is the kinematic viscosity and $f$ is forcing.

We generate a total of 200K training trajectories plus 1K testing trajectories using Exponax~\cite{koehler2024apebench}. 
We generate at a resolution of $256\times256$ with time step $\Delta t=1 \times 10^{-3}$ and simulate over $T=1$ seconds on a 2D square periodic domain of edge-length $L=2$. We store 10 frames in total and spectrally downsample the saved frames into resolution $64\times64$. 
The average per-trajectory simulation time is 0.0435 seconds.

\paragraph{1D, 2D and 3D Kuramoto-Sivashinsky}
We consider this equation on a periodic domain $\mathcal X=[0,L]^d$, where the spatial dimension is $d \in \{1, 2, 3\}$. While the 2D formulation serves as our standard benchmark, the 1D and 3D variants are included specifically for the difficulty scaling analysis. The governing equation for a scalar field $u(t,x)$ is:
\begin{align}
\label{eq:ks}
    \partial_t u \;+\; \Delta u \;+\; \Delta^2 u \;+\; \frac{1}{2}\|\nabla u\|_2^2 \;=\; 0,
\end{align}
where $\nabla$ and $\Delta$ denote the spatial gradient and Laplacian, respectively, and $\Delta^2$ is the biharmonic operator. 

We generate a total of 200K training trajectories plus 1K testing trajectories using Exponax~\cite{koehler2024apebench}.
We generate the data at resolution of 256 with time step $\Delta t=0.1$ and simulate over $T=10$ seconds on a 2D square periodic domain of edge-length $L=50$. We store 50 frames in total and spectrally downsample the saved frames into resolution $64\times64$.
The average per-trajectory simulation time for 1D is 0.0004 seconds.
The average per-trajectory simulation time for 2D is 0.0667 seconds.
The average per-trajectory simulation time for 3D is 2.7110 seconds.

\paragraph{2D Gray-Scott}
Let $u(t,x)$ and $v(t,x)$ denote chemical concentrations on a 2D domain $\mathcal X$.
The Gray--Scott system is formulated by
\begin{align}
    \partial_t u &= D_u \Delta u - u v^2 + F(1-u) \\
    \partial_t v &= D_v \Delta v + u v^2 - (F+k)\,v,
\end{align}
for diffusivities $D_u,D_v>0$ and reaction parameters $F,k$.

We generate a total of 200K training trajectories plus 1K testing trajectories using Exponax~\cite{koehler2024apebench}.
The reaction parameters are $F=0.029$ and $k=0.057$ and the diffusivities are set to $2.1 \times 10^{-5}$ and $1.1 \times 10^{-5}$.
We generate at a resolution of $256\times256$ with time step $\Delta t=0.5$ and simulate over $T=2000$ seconds on a 2D square periodic domain of edge-length $L=2$. 
The end-time is specifically chosen because most Gray-Scott simulations stabilize well before then.
We store 200 frames in total and spectrally downsample the saved frames into resolution of $64\times64$. 
The average per-trajectory simulation time is 0.4751 seconds.

\begin{figure}
\centering
\includegraphics[scale=0.7]{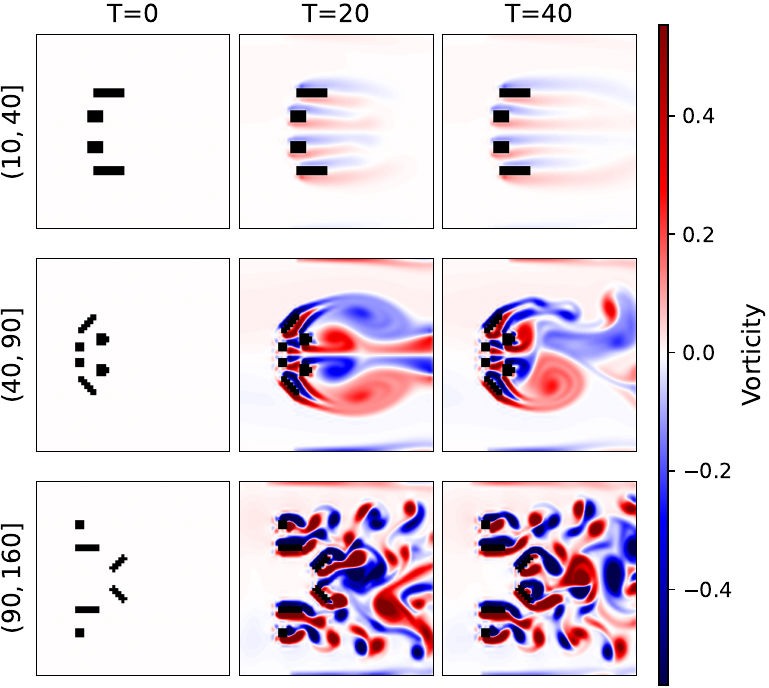}
\caption{Vorticity snapshots at $t=0,200,400$ of BreakFlow simulations of different $Re$ bins ($Re \in (10,40], (40,90]$ and $(90, 160]$) generated via PyFR.
There are inlet boundaries on all sides apart from an outlet on the right.
Black boxes represent obstacles.\looseness-1}
\label{fig:pyfr}
\end{figure}

\paragraph{BreakFlow}
The BreakFlow dataset is constructed to systematically evaluate the performance, scalability, and physical consistency of neural partial differential equation (PDE) solvers. An example of the resulting vorticity field is given in Fig.~\ref{fig:pyfr}.

Each simulation consists of four steps:
\begin{enumerate}
    \item Domain generation: we simulate over the square domain $[0,50]\times[-25,25]$ with an outlet boundary on the right and inlet boundaries on left, top, and bottom.
    We generate rectangular obstacles by looping in a nested fashion over the integers $x=\{11,\dots,25\}$ and $y\in\{0,\dots,15\}$;
    at each $(x,y)$ coordinate we place a rectangle with probability 0.5.
    The rectangle's dimensions are determined by randomly sampling an area $a$ between 4 and 30 (or 4 and 60 if $y=0$), then randomly sampling a height $h\in\mathbb Z$ and width $w\in\mathbb Z$ such that $wh=a$, $w\ge2$, and $h\ge2$.
    The rectangle's rotation is randomly sampled from $\{0,15,\dots,165\}$ degrees, except if $y=0$ in which case it is either not rotated or sampled from $\{0,45\}$ if it is a square ($h=w$).
    If the resulting rectangle overlaps or is closer than distance two to a previously placed rectangle, we do not place it.
    Lastly, to make the obstacles symmetric about the $y=0$ axis we take all rectangles with centers $(x,y)$ and place a mirrored rectangle at $(x,-y)$.
    \item Mesh generation: we generate a first-order triangular mesh using the Gmsh utility~\citep{gmsh}.
    The mesh is graded to have edge-length $\tfrac13$ near the obstacles, $\tfrac23$ at the edge of a refinement zone $[5,45]\times[-20,20]$, and 1 at the edge of the domain, with edge-sizes varying smoothly between them. 
    To decrease the simulation resolution, we simultaneously increase these mesh sizes.
    \item Simulation: we run PyFR using the settings of the 2D incompressible cylinder flow example,\footnote{\url{https://pyfr.readthedocs.io/en/latest/examples.html\#d-incompressible-cylinder-flow}} but extending the simulation time to 400.
    To decrease the simulation resolution, we increase the timestepping parameters {\tt dt} and {\tt pseudo-dt}.
    The average per-trajectory simulation time is 136.9045 seconds.
    \item Lastly, we convert the data from an unstructured mesh to a regular $64\times64$ grid using the ParaView utility~\citep{Ahrens2005ParaViewAE}.
    Both the gridded data and unstructured simulation files are available in the dataset repository.
\end{enumerate}

The dataset spans a Reynolds number range of $Re \in (10, 160]$. Data generation is organized into three discrete bins, with $Re$ sampled uniformly at random within each respective range: (1)~Bin 1: $Re \in (10, 40]$; (2)~Bin 2: $Re \in (40, 90]$ \, and (3)~Bin 3: $Re \in (90, 160]$. Each trajectory is simulated for 400 seconds, yielding 40 saved frames. This duration lets most trajectories reach limit-cycle behavior. For each Reynolds bin, we generate 2,000 training trajectories and 50 test trajectories. We preserve per-simulation wallclock time, velocity $x$, velocity $y$, pressure, and vorticity field in the benchmark dataset. All experiments are trained on $x$-velocity, $y$-velocity, and pressure fields, and report nRMSE averaged over these fields. The main experiments use shuffled combinations of all three bins for training and testing.

\section{Training Details}
\label{Adx:train}

We present the training settings used to compute breakeven complexity:
\begin{itemize}
    \item \textbf{FFNO} \cite{ffno}:
    FFNO is trained with model width of 64, 16 modes and 24 number of layers, learning rate of $1 \times 10^{-3}$ on periodic equations and $1 \times 10^{-4}$ on BreakFlow, and weight decay of $1 \times 10^{-4}$ across all equations. The batch size for training on periodic equations is 100, and is 16 for training on BreakFlow. The FFNO model has a total of 6,698,113 parameters.

    \item \textbf{EddyFormer} \cite{eddyformer}:
    EddyFormer is trained with model width of 32, FFN width of 128, 10 attention blocks, 8 attention heads, learning rate of $1 \times 10^{-3}$ on periodic equations and $1 \times 10^{-4}$ on BreakFlow, weight decay of $1 \times 10^{-4}$ across all equations. The batch size for training on periodic equations is 64, and is 16 for training on BreakFlow. The EddyFormer model has a total of 3,499,171 parameters.

    \item \textbf{DISCO} \cite{disco}: DISCO is trained with patch size 16, start channels 8, 4 down/4 up blocks, 4-group GroupNorm, 6 heads, 12 processor blocks, learning rate of $1 \times 10^{-3}$ on periodic equations and $1 \times 10^{-4}$ on BreakFlow, and weight decay of $1 \times 10^{-4}$ across all equations. The batch size for training on periodic equations is 64, and is 16 for training on BreakFlow. The DISCO model has a total of 100,765,696 parameters.

    \item \textbf{DPOT} \cite{dpot}: DPOT is trained with $8\times8$ patches, embedding width 512, 4 layers, 4 blocks, MLP ratio 1, output head dimension 32, Adam optimizer, learning rate of $1 \times 10^{-3}$ on periodic equations and $1 \times 10^{-4}$ on BreakFlow, and weight decay of $1 \times 10^{-4}$ across all equations. The batch size for training on periodic equations is 64, and is 16 for training on BreakFlow. The model has a total of 7,010,355 parameters.

    \item \textbf{Poseidon} \cite{poseidon}:
    Poseidon is tuned on pretrained Poseidon weights \footnote{https://huggingface.co/collections/camlab-ethz/poseidon}, with learning rate of $1 \times 10^{-3}$ on periodic equations and $1 \times 10^{-4}$ on BreakFlow, and weight decay of 0.01 across all equations. The batch size for Navier--Stokes is 100, for Kuramoto-Sivashinsky is 300, for Gray-Scott is 200, and for BreakFlow is 16. All Poseidon T, B, and L models share the same set of training settings. Poseidon T has 21M parameters. Poseidon B has 158M parameters. Poseidon L has 629M parameters.

    \item \textbf{HalfWalrus} \cite{walrus}: HalfWalrus is the $6.4\times10^8$-parameter variant introduced in \citet[Table~2]{walrus}, with the same model configuration. Its checkpoint was not public at the time of our experiments; we requested and received it from the Walrus team. We selected this variant after preliminary full-Walrus measurements gave $N^\star=\infty$, driven by its high inference cost, on all three periodic benchmarks.
    We tuned with learning rate of $1 \times 10^{-3}$, weight decay of 0.01 across all equations. The batch size for Navier--Stokes is 100, for Kuramoto--Sivashinsky is 300, for Gray-Scott is 200, and for BreakFlow is 16.
\end{itemize}

Training parameters are chosen to be the optimal values from parameter sweeps. Batch sizes and learning rates tend to be large to encourage faster convergence and lower up-front training cost. The reported budget $B$ includes data generation and the selected training run, excluding the parameter sweeps. 

For all training, we warm up for 5 training steps (including the forward pass, backward pass, optimizer step, etc.). We then run another five warm-up training steps, take the average of their step times as the estimate of training-step time, and use this estimate to reset the scheduler. The training budget is decremented using the measured wall-clock time during training.

All training uses the same hardware setup as data generation (Appendix~\ref{Adx:data}).

\paragraph{Inference timing.}
For each model--task pair, we use the largest inference batch that fits on a single NVIDIA L40S. We discard three complete warm-up inference steps, then average the batch runtime divided by batch size across inference batches to obtain $C_{\mathrm{inf}}$. Inference uses FP32 with no concurrent GPU jobs; this measures amortized time per trajectory for many independent solves.

\section{Data scaling fit procedure}
\label{Adx:scaling}

We use scaling laws as empirical models of the tradeoff between data generation and optimization, motivated by observed scaling behavior~\cite{ashton2025fluid,lu2021machine}. For a fixed model-task pair, we use a Chinchilla-style additive power-law model~\cite{hoffmann2022trainingcomputeoptimallargelanguage},
\begin{equation}
    \widehat{\mathcal L} (N_{\mathrm{data}},C_{\mathrm{train}}) = L_\infty + aN_{\mathrm{data}}^{-\alpha} + dC_{\mathrm{train}}^{-\beta},
    \label{eq:data_compute_scaling}
\end{equation}
where $N_{\mathrm{data}}$ is the number of training trajectories and $C_{\mathrm{train}}$ is the wall-clock time allocated to optimization. The coefficients $a,d>0$ control the magnitudes of the data and
optimization terms, while $\alpha,\beta>0$ are their respective scaling exponents. The fitted $L_\infty$ represents an effective error floor for the fixed model and training regime. For each model--task pair, we fit $(L_\infty,a,\alpha,d,\beta)$ independently to the measured average-case nRMSE of small-budget runs; these parameters are not assumed to be universal.

\paragraph{Predicted allocation frontier.}
Given a total budget $B$ and per-trajectory data-generation cost $C_{\mathrm{gen}}$, the allocation must satisfy $C_{\mathrm{gen}}N_{\mathrm{data}}+C_{\mathrm{train}}\le B$. Since the fitted error decreases with optimization time, the predicted allocation uses the remaining budget for training. We therefore define the budget-constrained error surface as
\begin{equation}
    \widehat{\mathcal L}_B(N_{\mathrm{data}}):=\widehat{\mathcal L} \bigl(N_{\mathrm{data}},B-C_{\mathrm{gen}}N_{\mathrm{data}} \bigr),
    \qquad
    0<N_{\mathrm{data}}<\frac{B}{C_{\mathrm{gen}}}.
    \label{eq:budget_scaling_surface}
\end{equation}
The predicted allocation frontier is
\begin{equation}
    \widehat N_{\mathrm{data}}(B)
    \in
    \operatorname*{arg\,min}_{
        0<N_{\mathrm{data}}<B/C_{\mathrm{gen}}
    }
    \widehat{\mathcal L}_B(N_{\mathrm{data}}).
    \label{eq:predicted_data_allocation}
\end{equation}
Here, $\widehat N_{\mathrm{data}}(B)$ denotes the predicted number of training trajectories, distinct from the breakeven complexity $N^\star$. This allocation minimizes the fitted average-case nRMSE; both average-case and worst-case breakeven complexities are subsequently evaluated for the same trained model using its measured errors.

\paragraph{Landscape and large-budget evaluation.}
For illustration, Fig.~\ref{fig:scaling} shows the fitted budget-constrained surface $\widehat{\mathcal L}_B(N_{\mathrm{data}})$ on a dense log-spaced grid with total-budget range $B\in[9\times10^2,7\times10^4]$ seconds and training-data range $N_{\mathrm{data}}\in[4\times10^2,3\times10^4]$. Feasible allocations satisfy $C_{\mathrm{gen}}N_{\mathrm{data}}<B$. The contour levels span the 3rd to 97th percentiles of the fitted values, with only a subset labeled for readability. The curve shows the predicted allocation frontier $\widehat N_{\mathrm{data}}(B)$, with small-budget markers at $B\in\{1000,2000,4000,8000\}$ seconds.

We train at the predicted large-budget allocations and report the measured errors of those models, rather than loss values read from the extrapolated surface. In particular, the 64K-second Poseidon-L allocation in Fig.~\ref{fig:scaling}
was predicted using only smaller-budget runs, and its displayed error was measured after training. This tests the achieved performance of the predicted configuration; it does not by itself establish globally optimal allocation or validate extrapolation for other models and tasks.

\section{Time vs. FLOP choice}
\label{Adx:time_flop_choice}

\begin{table}[!th]
\centering
\small
\begin{tabular}{lcc}
\toprule
Method / Workload & FLOP/sec & Relative to spectral solver \\
\midrule
FFNO training            & $3.542 \times 10^{11}$ & $90.40\times$ \\
FFNO inference           & $3.550 \times 10^{11}$ & $90.61\times$ \\
PyTorch spectral workload & $3.918 \times 10^{9}$  & $1.00\times$ \\
\bottomrule
\end{tabular}
\vspace{2mm}
\caption{Profiler-reported throughput (FLOP/sec) for warmed-up FP32 runs at batch size 100. All workloads use PyTorch and the same profiler; the reference is a representative RK4 pseudo-spectral workload, not the JAX/Exponax implementation used for the periodic benchmarks. Counts depend on profiler operation coverage.}
\label{tab:flops}
\end{table}

Although FLOPs are a common measure of computational cost in ML frameworks~\citep{hoffmann2022trainingcomputeoptimallargelanguage, abnar2025parameters}, when comparing neural and numerical PDE solvers FLOPs alone are not an adequate fair comparison standard, as the realized throughput (FLOP/sec) can differ substantially between workloads.

We profile FFNO training, FFNO inference, and a representative RK4 pseudo-spectral workload implemented in PyTorch on the same NVIDIA L40S GPU. JAX/XLA supports FLOP estimation, but its operation coverage and counting conventions differed from the PyTorch profiler in our tests. We therefore use a common profiler for this illustrative comparison and do not interpret it as a FLOP measurement of Exponax~\cite{koehler2024apebench}. Table~\ref{tab:flops} reports above $3.5\times10^{11}$ FLOP/sec for FFNO and $3.92\times10^9$ for the PyTorch spectral workload.

The reported ratios are approximately $90\times$, or nearly two orders of magnitude. They depend on workload structure and profiler coverage: operations delegated to optimized libraries, including FFTs, may be counted differently or incompletely. Low reported FLOP/sec does not by itself demonstrate low hardware utilization, especially for bandwidth-bound kernels. These numbers therefore motivate measuring elapsed cost directly; they do not establish an intrinsic utilization gap.
Moreover, many numerical methods depend strongly on CPU performance, memory access patterns, communication, and implementation details, making FLOP-based comparisons less fair across heterogeneous solver families. 

For the above reasons, and to be consistent with previous cost-aware research in the neural PDE solver community \cite{ashton2025fluid, weak_baseline}, we use wall-clock as the primary cost measure. Wall-clock can better reflect the actual tradeoff when choosing between a neural surrogate and an error-matched numerical solver.\looseness-1

\section{Interpreting breakeven complexity}
\label{Adx:interpretation}

Breakeven complexity ($N^\star$) is not an intrinsic property of a neural architecture or PDE. Rather, it is a deployment-dependent metric dictated by the chosen classical solver, error metric, hardware, and implementation details. Its value serves as a calibrated amortization threshold for a specific setup, not a universal ranking of solvers.

For $C_{\delta_B}>0$, define $r=C_{\mathrm{inf}}/C_{\delta_B}$. When $r<1$, Eq.~\eqref{eq:breakeven} becomes
\[
N^\star(B)=\frac{B}{C_{\delta_B}(1-r)}.
\]
A point estimate is informative when the positive cost gap is well resolved relative to timing variability. As $r$ approaches~1, small timing changes can produce large changes in $N^\star$; an uncertainty interval or a timing-sensitive flag is then more appropriate. If timing uncertainty spans $r=1$, both finite amortization and non-amortization remain possible. There is no universal numerical cutoff for $r$, since the useful threshold depends on the measurement uncertainty.
Timing uncertainty is distinct from variation due to training seeds or test trajectories.

Therefore, we treat breakeven complexity primarily as a cost-accounting tool. It highlights how training, inference, and baseline costs interact, exposing cases where accuracy-only evaluations are misleading. In practice, users can recalibrate Eq.~\eqref{eq:breakeven} for their specific target workloads and runtime environments.

\newpage
\section{Full Experimental Results Tables}
\label{Adx:table}

We present the full experimental data in Sec.~\ref{sec:eval} here. Tables~\ref{tab:results_full_ns}--\ref{tab:results_full_breakflow} show detailed results for Fig.~\ref{fig:result}. For budgets of 1K--8K seconds, we report the best configurations from parameter sweeps; for larger budgets, we train using the scaling-predicted number of trajectories. The scaling fit procedure is described in Appendix~\ref{Adx:scaling}. Data $\%$ denotes the percentage of budget allocated to data generation. Worst-case results are reported in \colorbox{blue!20}{blue cells}, while average-case results are reported in \colorbox{green!20}{green cells}.  Tables ~\ref{tab:result_dim}-- \ref{tab:poseidon_bin} show detailed results for Fig.~\ref{fig:difficulty}.

\begin{table*}[!htbp]
\centering
\small
\renewcommand{\arraystretch}{1.1000}
\setlength{\tabcolsep}{3pt}
\resizebox{\textwidth}{!}{%
\begin{tabular}{ll|cccc|cccc}
\toprule
\multicolumn{2}{c|}{\textbf{Navier-Stokes}} & \multicolumn{4}{c|}{\textbf{Supervised Model}} & \multicolumn{4}{c}{\textbf{Foundation Model}} \\
\hline
\textbf{Budget} & \textbf{Metric} & FFNO & EddyFormer & DPOT & DISCO & Poseidon T & Poseidon B & Poseidon L & HalfWalrus \\
\hline
\multirow{5}{*}{1e3} & Data \% & 12.19\% & 2.17\% & 17.42\% & 4.35\% & 9.05\% & 6.59\% & 5.07\% & 2.18\% \\
 & $\mathrm{nRMSE}_{\mathrm{worst}}$ & \cellcolor{blue!20}0.0472 & \cellcolor{blue!20}0.0490 & \cellcolor{blue!20}0.0942 & \cellcolor{blue!20}0.3520 & \cellcolor{blue!20}0.0161 & \cellcolor{blue!20}0.0168 & \cellcolor{blue!20}0.0326 & \cellcolor{blue!20}0.0497 \\
 & $\mathrm{nRMSE}_{\mathrm{avg}}$ & \cellcolor{green!20}0.0073 & \cellcolor{green!20}0.0093 & \cellcolor{green!20}0.0460 & \cellcolor{green!20}0.1955 & \cellcolor{green!20}0.0059 & \cellcolor{green!20}0.0077 & \cellcolor{green!20}0.0168 & \cellcolor{green!20}0.0127 \\
 & $N^\star_{\mathrm{worst}}$ & \cellcolor{blue!20}42717 & \cellcolor{blue!20}INF & \cellcolor{blue!20}46431 & \cellcolor{blue!20}INF & \cellcolor{blue!20}31014 & \cellcolor{blue!20}31226 & \cellcolor{blue!20}41960 & \cellcolor{blue!20}44122 \\
 & $N^\star_{\mathrm{avg}}$ & \cellcolor{green!20}32158 & \cellcolor{green!20}INF & \cellcolor{green!20}41661 & \cellcolor{green!20}1369566 & \cellcolor{green!20}27918 & \cellcolor{green!20}30668 & \cellcolor{green!20}38791 & \cellcolor{green!20}32950 \\
\hline
\multirow{5}{*}{2e3} & Data \% & 8.70\% & 1.74\% & 17.42\% & 4.35\% & 6.10\% & 4.61\% & 2.86\% & 2.18\% \\
 & $\mathrm{nRMSE}_{\mathrm{worst}}$ & \cellcolor{blue!20}0.0348 & \cellcolor{blue!20}0.0179 & \cellcolor{blue!20}0.0941 & \cellcolor{blue!20}0.3108 & \cellcolor{blue!20}0.0072 & \cellcolor{blue!20}0.0089 & \cellcolor{blue!20}0.0200 & \cellcolor{blue!20}0.0144 \\
 & $\mathrm{nRMSE}_{\mathrm{avg}}$ & \cellcolor{green!20}0.0056 & \cellcolor{green!20}0.0061 & \cellcolor{green!20}0.0436 & \cellcolor{green!20}0.1763 & \cellcolor{green!20}0.0030 & \cellcolor{green!20}0.0048 & \cellcolor{green!20}0.0102 & \cellcolor{green!20}0.0078 \\
 & $N^\star_{\mathrm{worst}}$ & \cellcolor{blue!20}85427 & \cellcolor{blue!20}INF & \cellcolor{blue!20}92862 & \cellcolor{blue!20}INF & \cellcolor{blue!20}55841 & \cellcolor{blue!20}62453 & \cellcolor{blue!20}77674 & \cellcolor{blue!20}65901 \\
 & $N^\star_{\mathrm{avg}}$ & \cellcolor{green!20}57689 & \cellcolor{green!20}10101000 & \cellcolor{green!20}83322 & \cellcolor{green!20}2739133 & \cellcolor{green!20}49450 & \cellcolor{green!20}53181 & \cellcolor{green!20}63397 & \cellcolor{green!20}58960 \\
\hline
\multirow{5}{*}{4e3} & Data \% & 6.53\% & 1.08\% & 17.42\% & 4.35\% & 3.36\% & 2.67\% & 1.54\% & 1.52\% \\
 & $\mathrm{nRMSE}_{\mathrm{worst}}$ & \cellcolor{blue!20}0.0257 & \cellcolor{blue!20}0.0166 & \cellcolor{blue!20}0.0876 & \cellcolor{blue!20}0.2821 & \cellcolor{blue!20}0.0040 & \cellcolor{blue!20}0.0058 & \cellcolor{blue!20}0.0096 & \cellcolor{blue!20}0.0058 \\
 & $\mathrm{nRMSE}_{\mathrm{avg}}$ & \cellcolor{green!20}0.0041 & \cellcolor{green!20}0.0041 & \cellcolor{green!20}0.0413 & \cellcolor{green!20}0.1469 & \cellcolor{green!20}0.0014 & \cellcolor{green!20}0.0026 & \cellcolor{green!20}0.0058 & \cellcolor{green!20}0.0037 \\
 & $N^\star_{\mathrm{worst}}$ & \cellcolor{blue!20}157929 & \cellcolor{blue!20}INF & \cellcolor{blue!20}185725 & \cellcolor{blue!20}INF & \cellcolor{blue!20}98907 & \cellcolor{blue!20}112369 & \cellcolor{blue!20}126918 & \cellcolor{blue!20}117920 \\
 & $N^\star_{\mathrm{avg}}$ & \cellcolor{green!20}101796 & \cellcolor{green!20}829222 & \cellcolor{green!20}166645 & \cellcolor{green!20}2311927 & \cellcolor{green!20}93240 & \cellcolor{green!20}93723 & \cellcolor{green!20}113896 & \cellcolor{green!20}103769 \\
\hline
\multirow{5}{*}{8e3} & Data \% & 4.35\% & 0.76\% & 10.88\% & 2.18\% & 1.52\% & 2.50\% & 0.89\% & 1.09\% \\
 & $\mathrm{nRMSE}_{\mathrm{worst}}$ & \cellcolor{blue!20}0.0219 & \cellcolor{blue!20}0.0155 & \cellcolor{blue!20}0.0797 & \cellcolor{blue!20}0.1505 & \cellcolor{blue!20}0.0023 & \cellcolor{blue!20}0.0027 & \cellcolor{blue!20}0.0050 & \cellcolor{blue!20}0.0055 \\
 & $\mathrm{nRMSE}_{\mathrm{avg}}$ & \cellcolor{green!20}0.0033 & \cellcolor{green!20}0.0026 & \cellcolor{green!20}0.0390 & \cellcolor{green!20}0.1203 & \cellcolor{green!20}0.0007 & \cellcolor{green!20}0.0010 & \cellcolor{green!20}0.0043 & \cellcolor{green!20}0.0034 \\
 & $N^\star_{\mathrm{worst}}$ & \cellcolor{blue!20}315858 & \cellcolor{blue!20}INF & \cellcolor{blue!20}369307 & \cellcolor{blue!20}10956532 & \cellcolor{blue!20}197800 & \cellcolor{blue!20}198888 & \cellcolor{blue!20}201436 & \cellcolor{blue!20}235840 \\
 & $N^\star_{\mathrm{avg}}$ & \cellcolor{green!20}203692 & \cellcolor{green!20}632117 & \cellcolor{green!20}331565 & \cellcolor{green!20}4623855 & \cellcolor{green!20}186481 & \cellcolor{green!20}187447 & \cellcolor{green!20}201281 & \cellcolor{green!20}207538 \\
\hline
\multirow{5}{*}{16e3} & Data \% & 3.18\% & 0.54\% & 6.80\% & 1.09\% & 0.93\% & 1.59\% & 0.49\% & 0.78\% \\
 & $\mathrm{nRMSE}_{\mathrm{worst}}$ & \cellcolor{blue!20}0.0162 & \cellcolor{blue!20}0.0090 & \cellcolor{blue!20}0.0769 & \cellcolor{blue!20}0.1427 & \cellcolor{blue!20}0.0011 & \cellcolor{blue!20}0.0016 & \cellcolor{blue!20}0.0027 & \cellcolor{blue!20}0.0042 \\
 & $\mathrm{nRMSE}_{\mathrm{avg}}$ & \cellcolor{green!20}0.0025 & \cellcolor{green!20}0.0017 & \cellcolor{green!20}0.0369 & \cellcolor{green!20}0.1145 & \cellcolor{green!20}0.0003 & \cellcolor{green!20}0.0004 & \cellcolor{green!20}0.0011 & \cellcolor{green!20}0.0022 \\
 & $N^\star_{\mathrm{worst}}$ & \cellcolor{blue!20}632197 & \cellcolor{blue!20}INF & \cellcolor{blue!20}738614 & \cellcolor{blue!20}21913064 & \cellcolor{blue!20}372962 & \cellcolor{blue!20}374895 & \cellcolor{blue!20}402554 & \cellcolor{blue!20}415077 \\
 & $N^\star_{\mathrm{avg}}$ & \cellcolor{green!20}407383 & \cellcolor{green!20}1264233 & \cellcolor{green!20}663130 & \cellcolor{green!20}9247710 & \cellcolor{green!20}372962 & \cellcolor{green!20}374895 & \cellcolor{green!20}379374 & \cellcolor{green!20}390225 \\
\hline
\multirow{5}{*}{32e3} & Data \% & 2.27\% & 0.37\% & 4.25\% & 0.55\% & 0.51\% & 1.13\% & 0.27\% & 0.56\% \\
 & $\mathrm{nRMSE}_{\mathrm{worst}}$ & \cellcolor{blue!20}0.0125 & \cellcolor{blue!20}0.0063 & \cellcolor{blue!20}0.0726 & \cellcolor{blue!20}0.1224 & \cellcolor{blue!20}0.0006 & \cellcolor{blue!20}0.0009 & \cellcolor{blue!20}0.0014 & \cellcolor{blue!20}0.0031 \\
 & $\mathrm{nRMSE}_{\mathrm{avg}}$ & \cellcolor{green!20}0.0019 & \cellcolor{green!20}0.0011 & \cellcolor{green!20}0.0350 & \cellcolor{green!20}0.0845 & \cellcolor{green!20}0.0001 & \cellcolor{green!20}0.0003 & \cellcolor{green!20}0.0007 & \cellcolor{green!20}0.0019 \\
 & $N^\star_{\mathrm{worst}}$ & \cellcolor{blue!20}1264395 & \cellcolor{blue!20}161611001 & \cellcolor{blue!20}1477229 & \cellcolor{blue!20}18495420 & \cellcolor{blue!20}745924 & \cellcolor{blue!20}749791 & \cellcolor{blue!20}758749 & \cellcolor{blue!20}780450 \\
 & $N^\star_{\mathrm{avg}}$ & \cellcolor{green!20}814768 & \cellcolor{green!20}1750351 & \cellcolor{green!20}1326260 & \cellcolor{green!20}7626332 & \cellcolor{green!20}745924 & \cellcolor{green!20}749791 & \cellcolor{green!20}758749 & \cellcolor{green!20}780450 \\
\hline
\multirow{5}{*}{64e3} & Data \% & 1.62\% & 0.26\% & 2.65\% & 0.28\% & 0.28\% & 0.80\% & 0.15\% & 0.40\% \\
 & $\mathrm{nRMSE}_{\mathrm{worst}}$ & \cellcolor{blue!20}0.0096 & \cellcolor{blue!20}0.0045 & \cellcolor{blue!20}0.0699 & \cellcolor{blue!20}0.1211 & \cellcolor{blue!20}0.0003 & \cellcolor{blue!20}0.0005 & \cellcolor{blue!20}0.0008 & \cellcolor{blue!20}0.0021 \\
 & $\mathrm{nRMSE}_{\mathrm{avg}}$ & \cellcolor{green!20}0.0014 & \cellcolor{green!20}0.0007 & \cellcolor{green!20}0.0331 & \cellcolor{green!20}0.0612 & \cellcolor{green!20}0.0001 & \cellcolor{green!20}0.0003 & \cellcolor{green!20}0.0004 & \cellcolor{green!20}0.0009 \\
 & $N^\star_{\mathrm{worst}}$ & \cellcolor{blue!20}2059435 & \cellcolor{blue!20}5056935 & \cellcolor{blue!20}2954459 & \cellcolor{blue!20}36990840 & \cellcolor{blue!20}1491848 & \cellcolor{blue!20}1499582 & \cellcolor{blue!20}1517499 & \cellcolor{blue!20}1560901 \\
 & $N^\star_{\mathrm{avg}}$ & \cellcolor{green!20}1629536 & \cellcolor{green!20}2586780 & \cellcolor{green!20}2652520 & \cellcolor{green!20}45252665 & \cellcolor{green!20}1491848 & \cellcolor{green!20}1499582 & \cellcolor{green!20}1517499 & \cellcolor{green!20}1560901 \\
\bottomrule
\end{tabular}%
}
\caption{Full results on Navier-Stokes for Fig.\ref{fig:result}. Worst-case results are shown in \colorbox{blue!20}{blue cells}, and average-case results are shown in \colorbox{green!20}{green cells}.}
\label{tab:results_full_ns}
\end{table*}

\begin{table*}[!htbp]
\centering
\small
\renewcommand{\arraystretch}{1.1000}
\setlength{\tabcolsep}{3pt}
\resizebox{\textwidth}{!}{%
\begin{tabular}{ll|cccc|cccc}
\toprule
\multicolumn{2}{c|}{\textbf{Kuramoto-Sivashinsky}} & \multicolumn{4}{c|}{\textbf{Supervised Model}} & \multicolumn{4}{c}{\textbf{Foundation Model}} \\
\hline
\textbf{Budget} & \textbf{Metric} & FFNO & EddyFormer & DPOT & DISCO & Poseidon T & Poseidon B & Poseidon L & HalfWalrus \\
\hline
\multirow{5}{*}{1e3} & Data \% & 6.67\% & 3.33\% & 6.67\% & 6.67\% & 5.66\% & 7.34\% & 13.35\% & 3.34\% \\
 & $\mathrm{nRMSE}_{\mathrm{worst}}$ & \cellcolor{blue!20}0.4228 & \cellcolor{blue!20}0.3451 & \cellcolor{blue!20}0.1499 & \cellcolor{blue!20}1.3481 & \cellcolor{blue!20}0.0330 & \cellcolor{blue!20}0.0353 & \cellcolor{blue!20}0.0561 & \cellcolor{blue!20}0.8980 \\
 & $\mathrm{nRMSE}_{\mathrm{avg}}$ & \cellcolor{green!20}0.1471 & \cellcolor{green!20}0.0236 & \cellcolor{green!20}0.0558 & \cellcolor{green!20}0.8650 & \cellcolor{green!20}0.0189 & \cellcolor{green!20}0.0182 & \cellcolor{green!20}0.0313 & \cellcolor{green!20}0.3699 \\
 & $N^\star_{\mathrm{worst}}$ & \cellcolor{blue!20}INF & \cellcolor{blue!20}INF & \cellcolor{blue!20}172131 & \cellcolor{blue!20}INF & \cellcolor{blue!20}46999 & \cellcolor{blue!20}47038 & \cellcolor{blue!20}49363 & \cellcolor{blue!20}INF \\
 & $N^\star_{\mathrm{avg}}$ & \cellcolor{green!20}533930 & \cellcolor{green!20}INF & \cellcolor{green!20}118618 & \cellcolor{green!20}INF & \cellcolor{green!20}45862 & \cellcolor{green!20}45875 & \cellcolor{green!20}50058 & \cellcolor{green!20}INF \\
\hline
\multirow{5}{*}{2e3} & Data \% & 4.67\% & 2.33\% & 13.35\% & 6.67\% & 4.88\% & 6.67\% & 11.34\% & 3.34\% \\
 & $\mathrm{nRMSE}_{\mathrm{worst}}$ & \cellcolor{blue!20}0.2617 & \cellcolor{blue!20}0.1684 & \cellcolor{blue!20}0.1148 & \cellcolor{blue!20}1.2241 & \cellcolor{blue!20}0.0252 & \cellcolor{blue!20}0.0254 & \cellcolor{blue!20}0.0319 & \cellcolor{blue!20}0.7438 \\
 & $\mathrm{nRMSE}_{\mathrm{avg}}$ & \cellcolor{green!20}0.0674 & \cellcolor{green!20}0.0169 & \cellcolor{green!20}0.0339 & \cellcolor{green!20}0.7581 & \cellcolor{green!20}0.0122 & \cellcolor{green!20}0.0123 & \cellcolor{green!20}0.0145 & \cellcolor{green!20}0.1837 \\
 & $N^\star_{\mathrm{worst}}$ & \cellcolor{blue!20}993589 & \cellcolor{blue!20}INF & \cellcolor{blue!20}332805 & \cellcolor{blue!20}INF & \cellcolor{blue!20}93998 & \cellcolor{blue!20}94077 & \cellcolor{blue!20}95143 & \cellcolor{blue!20}INF \\
 & $N^\star_{\mathrm{avg}}$ & \cellcolor{green!20}903789 & \cellcolor{green!20}INF & \cellcolor{green!20}190939 & \cellcolor{green!20}INF & \cellcolor{green!20}87702 & \cellcolor{green!20}87725 & \cellcolor{green!20}94082 & \cellcolor{green!20}INF \\
\hline
\multirow{5}{*}{4e3} & Data \% & 3.33\% & 1.66\% & 13.35\% & 13.35\% & 4.67\% & 6.34\% & 10.89\% & 2.34\% \\
 & $\mathrm{nRMSE}_{\mathrm{worst}}$ & \cellcolor{blue!20}0.1726 & \cellcolor{blue!20}0.1620 & \cellcolor{blue!20}0.1141 & \cellcolor{blue!20}0.9211 & \cellcolor{blue!20}0.0169 & \cellcolor{blue!20}0.0163 & \cellcolor{blue!20}0.0212 & \cellcolor{blue!20}0.1379 \\
 & $\mathrm{nRMSE}_{\mathrm{avg}}$ & \cellcolor{green!20}0.0349 & \cellcolor{green!20}0.0109 & \cellcolor{green!20}0.0189 & \cellcolor{green!20}0.6274 & \cellcolor{green!20}0.0075 & \cellcolor{green!20}0.0084 & \cellcolor{green!20}0.0107 & \cellcolor{green!20}0.0811 \\
 & $N^\star_{\mathrm{worst}}$ & \cellcolor{blue!20}863227 & \cellcolor{blue!20}INF & \cellcolor{blue!20}665610 & \cellcolor{blue!20}INF & \cellcolor{blue!20}187996 & \cellcolor{blue!20}188154 & \cellcolor{blue!20}197453 & \cellcolor{blue!20}INF \\
 & $N^\star_{\mathrm{avg}}$ & \cellcolor{green!20}598991 & \cellcolor{green!20}INF & \cellcolor{green!20}381879 & \cellcolor{green!20}INF & \cellcolor{green!20}175405 & \cellcolor{green!20}179759 & \cellcolor{green!20}188906 & \cellcolor{green!20}INF \\
\hline
\multirow{5}{*}{8e3} & Data \% & 2.33\% & 1.16\% & 1.67\% & 13.35\% & 4.58\% & 4.67\% & 6.67\% & 1.67\% \\
 & $\mathrm{nRMSE}_{\mathrm{worst}}$ & \cellcolor{blue!20}0.0641 & \cellcolor{blue!20}0.0518 & \cellcolor{blue!20}0.0561 & \cellcolor{blue!20}0.7674 & \cellcolor{blue!20}0.0103 & \cellcolor{blue!20}0.0097 & \cellcolor{blue!20}0.0183 & \cellcolor{blue!20}0.0695 \\
 & $\mathrm{nRMSE}_{\mathrm{avg}}$ & \cellcolor{green!20}0.0137 & \cellcolor{green!20}0.0063 & \cellcolor{green!20}0.0164 & \cellcolor{green!20}0.4775 & \cellcolor{green!20}0.0049 & \cellcolor{green!20}0.0049 & \cellcolor{green!20}0.0113 & \cellcolor{green!20}0.0463 \\
 & $N^\star_{\mathrm{worst}}$ & \cellcolor{blue!20}1726455 & \cellcolor{blue!20}INF & \cellcolor{blue!20}933011 & \cellcolor{blue!20}INF & \cellcolor{blue!20}375993 & \cellcolor{blue!20}376309 & \cellcolor{blue!20}383547 & \cellcolor{blue!20}INF \\
 & $N^\star_{\mathrm{avg}}$ & \cellcolor{green!20}1197982 & \cellcolor{green!20}INF & \cellcolor{green!20}753401 & \cellcolor{green!20}INF & \cellcolor{green!20}343967 & \cellcolor{green!20}344056 & \cellcolor{green!20}380574 & \cellcolor{green!20}INF \\
\hline
\multirow{5}{*}{16e3} & Data \% & 1.65\% & 0.82\% & 0.21\% & 13.35\% & 4.16\% & 4.34\% & 6.02\% & 1.19\% \\
 & $\mathrm{nRMSE}_{\mathrm{worst}}$ & \cellcolor{blue!20}0.0410 & \cellcolor{blue!20}0.0355 & \cellcolor{blue!20}0.0506 & \cellcolor{blue!20}0.7343 & \cellcolor{blue!20}0.0078 & \cellcolor{blue!20}0.0058 & \cellcolor{blue!20}0.0100 & \cellcolor{blue!20}0.0472 \\
 & $\mathrm{nRMSE}_{\mathrm{avg}}$ & \cellcolor{green!20}0.0067 & \cellcolor{green!20}0.0043 & \cellcolor{green!20}0.0143 & \cellcolor{green!20}0.4067 & \cellcolor{green!20}0.0035 & \cellcolor{green!20}0.0031 & \cellcolor{green!20}0.0073 & \cellcolor{green!20}0.0212 \\
 & $N^\star_{\mathrm{worst}}$ & \cellcolor{blue!20}3452906 & \cellcolor{blue!20}INF & \cellcolor{blue!20}1866020 & \cellcolor{blue!20}INF & \cellcolor{blue!20}718841 & \cellcolor{blue!20}688111 & \cellcolor{blue!20}761148 & \cellcolor{blue!20}INF \\
 & $N^\star_{\mathrm{avg}}$ & \cellcolor{green!20}2243547 & \cellcolor{green!20}INF & \cellcolor{green!20}1325361 & \cellcolor{green!20}INF & \cellcolor{green!20}687934 & \cellcolor{green!20}679346 & \cellcolor{green!20}726583 & \cellcolor{green!20}INF \\
\hline
\multirow{5}{*}{32e3} & Data \% & 1.16\% & 0.58\% & 0.03\% & 13.35\% & 3.89\% & 3.77\% & 4.87\% & 0.85\% \\
 & $\mathrm{nRMSE}_{\mathrm{worst}}$ & \cellcolor{blue!20}0.0223 & \cellcolor{blue!20}0.0200 & \cellcolor{blue!20}0.0437 & \cellcolor{blue!20}0.4432 & \cellcolor{blue!20}0.0060 & \cellcolor{blue!20}0.0032 & \cellcolor{blue!20}0.0079 & \cellcolor{blue!20}0.0311 \\
 & $\mathrm{nRMSE}_{\mathrm{avg}}$ & \cellcolor{green!20}0.0031 & \cellcolor{green!20}0.0028 & \cellcolor{green!20}0.0113 & \cellcolor{green!20}0.3201 & \cellcolor{green!20}0.0024 & \cellcolor{green!20}0.0019 & \cellcolor{green!20}0.0048 & \cellcolor{green!20}0.0124 \\
 & $N^\star_{\mathrm{worst}}$ & \cellcolor{blue!20}5680026 & \cellcolor{blue!20}INF & \cellcolor{blue!20}3732041 & \cellcolor{blue!20}INF & \cellcolor{blue!20}1375868 & \cellcolor{blue!20}1358347 & \cellcolor{blue!20}1453166 & \cellcolor{blue!20}INF \\
 & $N^\star_{\mathrm{avg}}$ & \cellcolor{green!20}3935282 & \cellcolor{green!20}INF & \cellcolor{green!20}2650722 & \cellcolor{green!20}INF & \cellcolor{green!20}1342382 & \cellcolor{green!20}1342721 & \cellcolor{green!20}1423833 & \cellcolor{green!20}INF \\
\hline
\multirow{5}{*}{64e3} & Data \% & 0.82\% & 0.41\% & 0.00\% & 13.35\% & 3.63\% & 3.28\% & 3.94\% & 0.61\% \\
 & $\mathrm{nRMSE}_{\mathrm{worst}}$ & \cellcolor{blue!20}0.0122 & \cellcolor{blue!20}0.0113 & \cellcolor{blue!20}0.0303 & \cellcolor{blue!20}0.4017 & \cellcolor{blue!20}0.0029 & \cellcolor{blue!20}0.0031 & \cellcolor{blue!20}0.0052 & \cellcolor{blue!20}0.0142 \\
 & $\mathrm{nRMSE}_{\mathrm{avg}}$ & \cellcolor{green!20}0.0018 & \cellcolor{green!20}0.0018 & \cellcolor{green!20}0.0099 & \cellcolor{green!20}0.2941 & \cellcolor{green!20}0.0016 & \cellcolor{green!20}0.0016 & \cellcolor{green!20}0.0035 & \cellcolor{green!20}0.0075 \\
 & $N^\star_{\mathrm{worst}}$ & \cellcolor{blue!20}9583855 & \cellcolor{blue!20}INF & \cellcolor{blue!20}6027208 & \cellcolor{blue!20}INF & \cellcolor{blue!20}2751736 & \cellcolor{blue!20}2717387 & \cellcolor{blue!20}2847666 & \cellcolor{blue!20}INF \\
 & $N^\star_{\mathrm{avg}}$ & \cellcolor{green!20}7346417 & \cellcolor{green!20}INF & \cellcolor{green!20}5301444 & \cellcolor{green!20}INF & \cellcolor{green!20}2684765 & \cellcolor{green!20}2685443 & \cellcolor{green!20}2810154 & \cellcolor{green!20}INF \\
\bottomrule
\end{tabular}%
}
\caption{Full results on Kuramoto-Sivashinsky for Fig.\ref{fig:result}. Worst-case results are shown in \colorbox{blue!20}{blue cells}, and average-case results are shown in \colorbox{green!20}{green cells}.}
\label{tab:results_full_ks}
\end{table*}

\begin{table*}[!htbp]
\centering
\small
\renewcommand{\arraystretch}{1.1000}
\setlength{\tabcolsep}{3pt}
\resizebox{\textwidth}{!}{%
\begin{tabular}{ll|cccc|cccc}
\toprule
\multicolumn{2}{c|}{\textbf{Gray-Scott}} & \multicolumn{4}{c|}{\textbf{Supervised Model}} & \multicolumn{4}{c}{\textbf{Foundation Model}} \\
\hline
\textbf{Budget} & \textbf{Metric} & FFNO & EddyFormer & DPOT & DISCO & Poseidon T & Poseidon B & Poseidon L & HalfWalrus \\
\hline
\multirow{5}{*}{1e3} & Data \% & 23.75\% & 11.87\% & 23.75\% & 23.75\% & 31.73\% & 33.39\% & 32.96\% & 47.51\% \\
 & $\mathrm{nRMSE}_{\mathrm{worst}}$ & \cellcolor{blue!20}0.6393 & \cellcolor{blue!20}0.6444 & \cellcolor{blue!20}0.4040 & \cellcolor{blue!20}12.7105 & \cellcolor{blue!20}0.0361 & \cellcolor{blue!20}0.0364 & \cellcolor{blue!20}0.0561 & \cellcolor{blue!20}0.8677 \\
 & $\mathrm{nRMSE}_{\mathrm{avg}}$ & \cellcolor{green!20}0.3432 & \cellcolor{green!20}0.3669 & \cellcolor{green!20}0.2988 & \cellcolor{green!20}3.5880 & \cellcolor{green!20}0.0075 & \cellcolor{green!20}0.0106 & \cellcolor{green!20}0.0147 & \cellcolor{green!20}0.5164 \\
 & $N^\star_{\mathrm{worst}}$ & \cellcolor{blue!20}43146 & \cellcolor{blue!20}33962 & \cellcolor{blue!20}INF & \cellcolor{blue!20}INF & \cellcolor{blue!20}4479 & \cellcolor{blue!20}4524 & \cellcolor{blue!20}4625 & \cellcolor{blue!20}INF \\
 & $N^\star_{\mathrm{avg}}$ & \cellcolor{green!20}34271 & \cellcolor{green!20}28212 & \cellcolor{green!20}576238 & \cellcolor{green!20}INF & \cellcolor{green!20}2614 & \cellcolor{green!20}2664 & \cellcolor{green!20}3643 & \cellcolor{green!20}INF \\
\hline
\multirow{5}{*}{2e3} & Data \% & 16.62\% & 17.81\% & 23.75\% & 23.75\% & 21.06\% & 19.53\% & 21.63\% & 47.51\% \\
 & $\mathrm{nRMSE}_{\mathrm{worst}}$ & \cellcolor{blue!20}0.5674 & \cellcolor{blue!20}0.4790 & \cellcolor{blue!20}0.3829 & \cellcolor{blue!20}7.4014 & \cellcolor{blue!20}0.0250 & \cellcolor{blue!20}0.0260 & \cellcolor{blue!20}0.0334 & \cellcolor{blue!20}0.7692 \\
 & $\mathrm{nRMSE}_{\mathrm{avg}}$ & \cellcolor{green!20}0.2179 & \cellcolor{green!20}0.2145 & \cellcolor{green!20}0.2753 & \cellcolor{green!20}1.6723 & \cellcolor{green!20}0.0051 & \cellcolor{green!20}0.0082 & \cellcolor{green!20}0.0106 & \cellcolor{green!20}0.4752 \\
 & $N^\star_{\mathrm{worst}}$ & \cellcolor{blue!20}54918 & \cellcolor{blue!20}46312 & \cellcolor{blue!20}INF & \cellcolor{blue!20}INF & \cellcolor{blue!20}5668 & \cellcolor{blue!20}5704 & \cellcolor{blue!20}6790 & \cellcolor{blue!20}INF \\
 & $N^\star_{\mathrm{avg}}$ & \cellcolor{green!20}35571 & \cellcolor{green!20}32003 & \cellcolor{green!20}731156 & \cellcolor{green!20}INF & \cellcolor{green!20}5294 & \cellcolor{green!20}5545 & \cellcolor{green!20}5785 & \cellcolor{green!20}INF \\
\hline
\multirow{5}{*}{4e3} & Data \% & 9.72\% & 11.87\% & 23.75\% & 23.75\% & 14.55\% & 13.71\% & 16.07\% & 47.51\% \\
 & $\mathrm{nRMSE}_{\mathrm{worst}}$ & \cellcolor{blue!20}0.2592 & \cellcolor{blue!20}0.1489 & \cellcolor{blue!20}0.3459 & \cellcolor{blue!20}2.4451 & \cellcolor{blue!20}0.0206 & \cellcolor{blue!20}0.0181 & \cellcolor{blue!20}0.0231 & \cellcolor{blue!20}0.7050 \\
 & $\mathrm{nRMSE}_{\mathrm{avg}}$ & \cellcolor{green!20}0.1571 & \cellcolor{green!20}0.0724 & \cellcolor{green!20}0.2488 & \cellcolor{green!20}0.9755 & \cellcolor{green!20}0.0047 & \cellcolor{green!20}0.0061 & \cellcolor{green!20}0.0080 & \cellcolor{green!20}0.4224 \\
 & $N^\star_{\mathrm{worst}}$ & \cellcolor{blue!20}57810 & \cellcolor{blue!20}43135 & \cellcolor{blue!20}INF & \cellcolor{blue!20}INF & \cellcolor{blue!20}11337 & \cellcolor{blue!20}11409 & \cellcolor{blue!20}11610 & \cellcolor{blue!20}INF \\
 & $N^\star_{\mathrm{avg}}$ & \cellcolor{green!20}46262 & \cellcolor{green!20}38150 & \cellcolor{green!20}1462313 & \cellcolor{green!20}INF & \cellcolor{green!20}10850 & \cellcolor{green!20}11091 & \cellcolor{green!20}11570 & \cellcolor{green!20}INF \\
\hline
\multirow{5}{*}{8e3} & Data \% & 5.93\% & 8.31\% & 23.75\% & 23.75\% & 8.67\% & 8.10\% & 8.63\% & 33.36\% \\
 & $\mathrm{nRMSE}_{\mathrm{worst}}$ & \cellcolor{blue!20}0.2451 & \cellcolor{blue!20}0.1188 & \cellcolor{blue!20}0.3451 & \cellcolor{blue!20}2.2996 & \cellcolor{blue!20}0.0149 & \cellcolor{blue!20}0.0151 & \cellcolor{blue!20}0.0161 & \cellcolor{blue!20}0.5446 \\
 & $\mathrm{nRMSE}_{\mathrm{avg}}$ & \cellcolor{green!20}0.1168 & \cellcolor{green!20}0.0556 & \cellcolor{green!20}0.2319 & \cellcolor{green!20}0.7551 & \cellcolor{green!20}0.0032 & \cellcolor{green!20}0.0053 & \cellcolor{green!20}0.0056 & \cellcolor{green!20}0.3122 \\
 & $N^\star_{\mathrm{worst}}$ & \cellcolor{blue!20}115620 & \cellcolor{blue!20}76301 & \cellcolor{blue!20}16127867 & \cellcolor{blue!20}INF & \cellcolor{blue!20}22675 & \cellcolor{blue!20}22819 & \cellcolor{blue!20}22487 & \cellcolor{blue!20}INF \\
 & $N^\star_{\mathrm{avg}}$ & \cellcolor{green!20}92524 & \cellcolor{green!20}59907 & \cellcolor{green!20}2924626 & \cellcolor{green!20}INF & \cellcolor{green!20}22178 & \cellcolor{green!20}22182 & \cellcolor{green!20}22196 & \cellcolor{green!20}INF \\
\hline
\multirow{5}{*}{16e3} & Data \% & 3.79\% & 8.31\% & 23.75\% & 23.75\% & 6.89\% & 5.19\% & 6.03\% & 23.75\% \\
 & $\mathrm{nRMSE}_{\mathrm{worst}}$ & \cellcolor{blue!20}0.1561 & \cellcolor{blue!20}0.0571 & \cellcolor{blue!20}0.3297 & \cellcolor{blue!20}1.1566 & \cellcolor{blue!20}0.0119 & \cellcolor{blue!20}0.0114 & \cellcolor{blue!20}0.0126 & \cellcolor{blue!20}0.4909 \\
 & $\mathrm{nRMSE}_{\mathrm{avg}}$ & \cellcolor{green!20}0.0790 & \cellcolor{green!20}0.0247 & \cellcolor{green!20}0.2177 & \cellcolor{green!20}0.6737 & \cellcolor{green!20}0.0026 & \cellcolor{green!20}0.0036 & \cellcolor{green!20}0.0031 & \cellcolor{green!20}0.2823 \\
 & $N^\star_{\mathrm{worst}}$ & \cellcolor{blue!20}185048 & \cellcolor{blue!20}119813 & \cellcolor{blue!20}5849253 & \cellcolor{blue!20}INF & \cellcolor{blue!20}45394 & \cellcolor{blue!20}45632 & \cellcolor{blue!20}46264 & \cellcolor{blue!20}INF \\
 & $N^\star_{\mathrm{avg}}$ & \cellcolor{green!20}138107 & \cellcolor{green!20}91481 & \cellcolor{green!20}709758 & \cellcolor{green!20}INF & \cellcolor{green!20}41900 & \cellcolor{green!20}42630 & \cellcolor{green!20}43181 & \cellcolor{green!20}INF \\
\hline
\multirow{5}{*}{32e3} & Data \% & 2.37\% & 7.17\% & 23.75\% & 23.75\% & 4.72\% & 3.28\% & 3.91\% & 16.45\% \\
 & $\mathrm{nRMSE}_{\mathrm{worst}}$ & \cellcolor{blue!20}0.1082 & \cellcolor{blue!20}0.0306 & \cellcolor{blue!20}0.3276 & \cellcolor{blue!20}1.2234 & \cellcolor{blue!20}0.0083 & \cellcolor{blue!20}0.0081 & \cellcolor{blue!20}0.0072 & \cellcolor{blue!20}0.4232 \\
 & $\mathrm{nRMSE}_{\mathrm{avg}}$ & \cellcolor{green!20}0.0554 & \cellcolor{green!20}0.0126 & \cellcolor{green!20}0.2134 & \cellcolor{green!20}0.6279 & \cellcolor{green!20}0.0017 & \cellcolor{green!20}0.0018 & \cellcolor{green!20}0.0028 & \cellcolor{green!20}0.2399 \\
 & $N^\star_{\mathrm{worst}}$ & \cellcolor{blue!20}308477 & \cellcolor{blue!20}182962 & \cellcolor{blue!20}11698506 & \cellcolor{blue!20}INF & \cellcolor{blue!20}90788 & \cellcolor{blue!20}91264 & \cellcolor{blue!20}92528 & \cellcolor{blue!20}INF \\
 & $N^\star_{\mathrm{avg}}$ & \cellcolor{green!20}221385 & \cellcolor{green!20}101407 & \cellcolor{green!20}1419517 & \cellcolor{green!20}INF & \cellcolor{green!20}67869 & \cellcolor{green!20}68135 & \cellcolor{green!20}85280 & \cellcolor{green!20}INF \\
\hline
\multirow{5}{*}{64e3} & Data \% & 1.48\% & 6.19\% & 23.75\% & 23.75\% & 3.24\% & 2.07\% & 2.54\% & 11.55\% \\
 & $\mathrm{nRMSE}_{\mathrm{worst}}$ & \cellcolor{blue!20}0.0751 & \cellcolor{blue!20}0.0164 & \cellcolor{blue!20}0.3177 & \cellcolor{blue!20}0.9675 & \cellcolor{blue!20}0.0068 & \cellcolor{blue!20}0.0064 & \cellcolor{blue!20}0.0048 & \cellcolor{blue!20}0.3849 \\
 & $\mathrm{nRMSE}_{\mathrm{avg}}$ & \cellcolor{green!20}0.0388 & \cellcolor{green!20}0.0064 & \cellcolor{green!20}0.2020 & \cellcolor{green!20}0.5895 & \cellcolor{green!20}0.0015 & \cellcolor{green!20}0.0010 & \cellcolor{green!20}0.0018 & \cellcolor{green!20}0.2060 \\
 & $N^\star_{\mathrm{worst}}$ & \cellcolor{blue!20}552428 & \cellcolor{blue!20}198286 & \cellcolor{blue!20}23397013 & \cellcolor{blue!20}INF & \cellcolor{blue!20}169688 & \cellcolor{blue!20}170520 & \cellcolor{blue!20}172724 & \cellcolor{blue!20}INF \\
 & $N^\star_{\mathrm{avg}}$ & \cellcolor{green!20}396305 & \cellcolor{green!20}188096 & \cellcolor{green!20}2839035 & \cellcolor{green!20}INF & \cellcolor{green!20}135739 & \cellcolor{green!20}136271 & \cellcolor{green!20}137674 & \cellcolor{green!20}INF \\
\bottomrule
\end{tabular}%
}
\caption{Full results on Gray-Scott for Fig.\ref{fig:result}. Worst-case results are shown in \colorbox{blue!20}{blue cells}, and average-case results are shown in \colorbox{green!20}{green cells}.}
\label{tab:results_full_gs}
\end{table*}

\begin{table*}[!htbp]
\centering
\small
\renewcommand{\arraystretch}{1.1000}
\setlength{\tabcolsep}{3pt}
\resizebox{\textwidth}{!}{%
\begin{tabular}{ll|cccc|cccc}
\toprule
\multicolumn{2}{c|}{\textbf{BreakFlow}} & \multicolumn{4}{c|}{\textbf{Supervised Model}} & \multicolumn{4}{c}{\textbf{Foundation Model}} \\
\hline
\textbf{Budget} & \textbf{Metric} & FFNO & EddyFormer & DPOT & DISCO & Poseidon T & Poseidon B & Poseidon L & HalfWalrus \\
\hline
\multirow{7}{*}{16e4} & Data \% & 64.17\% & 64.17\% & 64.17\% & 64.17\% & 85.56\% & 85.56\% & 85.56\% & 64.17\% \\
 & $\mathrm{nRMSE}_{\mathrm{worst}}$ & \cellcolor{blue!20}5.9541 & \cellcolor{blue!20}1.2171 & \cellcolor{blue!20}2.1317 & \cellcolor{blue!20}3.0537 & \cellcolor{blue!20}0.6455 & \cellcolor{blue!20}0.5411 & \cellcolor{blue!20}0.5009 & \cellcolor{blue!20}1.0234 \\
 & $\mathrm{nRMSE}_{\mathrm{avg}}$ & \cellcolor{green!20}0.1524 & \cellcolor{green!20}0.4245 & \cellcolor{green!20}0.0642 & \cellcolor{green!20}1.3549 & \cellcolor{green!20}0.0853 & \cellcolor{green!20}0.0691 & \cellcolor{green!20}0.0694 & \cellcolor{green!20}0.2118 \\
 & $N^\star_{\mathrm{worst}}$ & \cellcolor{blue!20}10695 & \cellcolor{blue!20}10692 & \cellcolor{blue!20}10673 & \cellcolor{blue!20}12053 & \cellcolor{blue!20}10686 & \cellcolor{blue!20}10700 & \cellcolor{blue!20}10700 & \cellcolor{blue!20}11809 \\
 & $N^\star_{\mathrm{avg}}$ & \cellcolor{green!20}1991 & \cellcolor{green!20}10240 & \cellcolor{green!20}1327 & \cellcolor{green!20}12053 & \cellcolor{green!20}1327 & \cellcolor{green!20}1327 & \cellcolor{green!20}1328 & \cellcolor{green!20}5841 \\
\hline
\multirow{5}{*}{25e4} & Data \% & 54.76\% & 54.76\% & 54.76\% & 54.76\% & 68.45\% & 68.45\% & 82.14\% & 68.45\% \\
 & $\mathrm{nRMSE}_{\mathrm{worst}}$ & \cellcolor{blue!20}5.7931 & \cellcolor{blue!20}0.9283 & \cellcolor{blue!20}1.9345 & \cellcolor{blue!20}2.8313 & \cellcolor{blue!20}0.5856 & \cellcolor{blue!20}0.5271 & \cellcolor{blue!20}0.4762 & \cellcolor{blue!20}0.9798 \\
 & $\mathrm{nRMSE}_{\mathrm{avg}}$ & \cellcolor{green!20}0.1430 & \cellcolor{green!20}0.3490 & \cellcolor{green!20}0.0569 & \cellcolor{green!20}1.1879 & \cellcolor{green!20}0.0798 & \cellcolor{green!20}0.0653 & \cellcolor{green!20}0.0630 & \cellcolor{green!20}0.1836 \\
 & $N^\star_{\mathrm{worst}}$ & \cellcolor{blue!20}16711 & \cellcolor{blue!20}16707 & \cellcolor{blue!20}16677 & \cellcolor{blue!20}18833 & \cellcolor{blue!20}16697 & \cellcolor{blue!20}16719 & \cellcolor{blue!20}16724 & \cellcolor{blue!20}18451 \\
 & $N^\star_{\mathrm{avg}}$ & \cellcolor{green!20}2139 & \cellcolor{green!20}16001 & \cellcolor{green!20}1826 & \cellcolor{green!20}18833 & \cellcolor{green!20}1826 & \cellcolor{green!20}1827 & \cellcolor{green!20}1827 & \cellcolor{green!20}5626 \\
\hline
\multirow{7}{*}{33e4} & Data \% & 41.48\% & 41.48\% & 62.22\% & 62.22\% & 62.22\% & 62.22\% & 72.60\% & 62.22\% \\
 & $\mathrm{nRMSE}_{\mathrm{worst}}$ & \cellcolor{blue!20}5.6499 & \cellcolor{blue!20}0.6715 & \cellcolor{blue!20}1.7593 & \cellcolor{blue!20}2.6337 & \cellcolor{blue!20}0.5324 & \cellcolor{blue!20}0.5146 & \cellcolor{blue!20}0.4543 & \cellcolor{blue!20}0.9410 \\
 & $\mathrm{nRMSE}_{\mathrm{avg}}$ & \cellcolor{green!20}0.1356 & \cellcolor{green!20}0.2818 & \cellcolor{green!20}0.0504 & \cellcolor{green!20}1.0394 & \cellcolor{green!20}0.0749 & \cellcolor{green!20}0.0619 & \cellcolor{green!20}0.0574 & \cellcolor{green!20}0.1787 \\
 & $N^\star_{\mathrm{worst}}$ & \cellcolor{blue!20}22058 & \cellcolor{blue!20}22053 & \cellcolor{blue!20}22014 & \cellcolor{blue!20}22053 & \cellcolor{blue!20}22041 & \cellcolor{blue!20}22069 & \cellcolor{blue!20}22070 & \cellcolor{blue!20}24356 \\
 & $N^\star_{\mathrm{avg}}$ & \cellcolor{green!20}3096 & \cellcolor{green!20}18746 & \cellcolor{green!20}2737 & \cellcolor{green!20}22053 & \cellcolor{green!20}2737 & \cellcolor{green!20}2737 & \cellcolor{green!20}2738 & \cellcolor{green!20}5453 \\
\hline
\multirow{7}{*}{42e4} & Data \% & 40.74\% & 40.74\% & 57.04\% & 57.04\% & 57.04\% & 57.04\% & 65.19\% & 57.04\% \\
 & $\mathrm{nRMSE}_{\mathrm{worst}}$ & \cellcolor{blue!20}5.4479 & \cellcolor{blue!20}0.2712 & \cellcolor{blue!20}1.3122 & \cellcolor{blue!20}1.4679 & \cellcolor{blue!20}0.5137 & \cellcolor{blue!20}0.5146 & \cellcolor{blue!20}0.4408 & \cellcolor{blue!20}0.8368 \\
 & $\mathrm{nRMSE}_{\mathrm{avg}}$ & \cellcolor{green!20}0.1121 & \cellcolor{green!20}0.1586 & \cellcolor{green!20}0.0410 & \cellcolor{green!20}0.8147 & \cellcolor{green!20}0.0695 & \cellcolor{green!20}0.0544 & \cellcolor{green!20}0.0526 & \cellcolor{green!20}0.1594 \\
 & $N^\star_{\mathrm{worst}}$ & \cellcolor{blue!20}28074 & \cellcolor{blue!20}28068 & \cellcolor{blue!20}28018 & \cellcolor{blue!20}28068 & \cellcolor{blue!20}28052 & \cellcolor{blue!20}28088 & \cellcolor{blue!20}28089 & \cellcolor{blue!20}30998 \\
 & $N^\star_{\mathrm{avg}}$ & \cellcolor{green!20}3068 & \cellcolor{green!20}5227 & \cellcolor{green!20}3483 & \cellcolor{green!20}28068 & \cellcolor{green!20}3483 & \cellcolor{green!20}3483 & \cellcolor{green!20}3484 & \cellcolor{green!20}5321 \\
\bottomrule
\end{tabular}%
}
\caption{Full nRMSE and breakeven-complexity results on BreakFlow for Fig.\ref{fig:result}. Worst-case results are shown in \colorbox{blue!20}{blue cells}, and average-case results are shown in \colorbox{green!20}{green cells}. 
For BreakFlow, the larger average--worst gap may be related to flow separation around multiple obstacles, nonperiodic boundaries, and substantial variation in difficulty across Reynolds-number bins; differences in vortex timing or phase can also produce large pointwise nRMSE. Limited training-data coverage due to expensive simulation may further contribute to this gap, although these are plausible explanations rather than isolated causal findings.
}
\label{tab:results_full_breakflow}
\end{table*}

\begin{table*}[!htbp]
\centering
\small
\renewcommand{\arraystretch}{1.10}
\setlength{\tabcolsep}{5pt}
\resizebox{0.92\textwidth}{!}{%
\begin{tabular}{c|cc|cc|cc}
\toprule
\multirow{2}{*}{\textbf{Budget}} 
& \multicolumn{2}{c|}{\textbf{1D Kuramoto--Sivashinsky}} 
& \multicolumn{2}{c|}{\textbf{2D Kuramoto--Sivashinsky}} 
& \multicolumn{2}{c}{\textbf{3D Kuramoto--Sivashinsky}} \\
\cline{2-7}
& $\mathrm{nRMSE}_{\mathrm{avg}}$ & $N^\star_{\mathrm{avg}}$
& $\mathrm{nRMSE}_{\mathrm{avg}}$ & $N^\star_{\mathrm{avg}}$
& $\mathrm{nRMSE}_{\mathrm{avg}}$ & $N^\star_{\mathrm{avg}}$ \\
\hline
$1e3$  & 0.002345 & INF & 0.1471 & 533930  & 0.4886 & INF \\
$2e3$  & 0.002224 & INF & 0.0674 & 903789  & 0.1212 & INF \\
$4e3$  & 0.002164 & INF & 0.0349 & 598991  & 0.0599 & 6531 \\
$8e3$  & 0.002112 & INF & 0.0137 & 1197982 & 0.0305 & 14799 \\
$16e3$ & 0.002030 & INF & 0.0067 & 2243547 & 0.0107 & 29637 \\
$32e3$ & 0.001964 & INF & 0.0031 & 3935282 & 0.0043 & 59274 \\
$64e3$ & 0.001715 & INF & 0.0018 & 7346417 & 0.0020 & 94378 \\
\bottomrule
\end{tabular}%
}
\caption{FFNO results across PDE dimensionality for Kuramoto--Sivashinsky equation for Fig.\ref{fig:difficulty}.}
\label{tab:result_dim}
\end{table*}

\begin{table*}[!htbp]
\centering
\small
\renewcommand{\arraystretch}{1.10}
\setlength{\tabcolsep}{5pt}
\resizebox{\textwidth}{!}{%
\begin{tabular}{c|cc|cc|cc|cc}
\toprule
\multirow{2}{*}{\textbf{Budget}} 
& \multicolumn{2}{c|}{\textbf{Rollout length = 25}} 
& \multicolumn{2}{c|}{\textbf{Rollout length = 50}} 
& \multicolumn{2}{c|}{\textbf{Rollout length = 100}} 
& \multicolumn{2}{c}{\textbf{Rollout length = 200}} \\
\cline{2-9}
& $\mathrm{nRMSE}_{\mathrm{avg}}$ & $N^\star_{\mathrm{avg}}$
& $\mathrm{nRMSE}_{\mathrm{avg}}$ & $N^\star_{\mathrm{avg}}$
& $\mathrm{nRMSE}_{\mathrm{avg}}$ & $N^\star_{\mathrm{avg}}$
& $\mathrm{nRMSE}_{\mathrm{avg}}$ & $N^\star_{\mathrm{avg}}$ \\
\hline
$1e3$  & 0.0413 & 73188   & 0.1189 & 42804   & 0.1947 & 33964  & 0.3432 & 34271 \\
$2e3$  & 0.0262 & 134418  & 0.0704 & 75827   & 0.1408 & 47072  & 0.2179 & 35571 \\
$4e3$  & 0.0149 & 229795  & 0.0462 & 132580  & 0.0791 & 69130  & 0.1571 & 46262 \\
$8e3$  & 0.0121 & 432644  & 0.0342 & 238220  & 0.0583 & 125464 & 0.1168 & 92524 \\
$16e3$ & 0.0073 & 721022  & 0.0213 & 389005  & 0.0371 & 204212 & 0.0790 & 138107 \\
$32e3$ & 0.0048 & 1266775 & 0.0141 & 641305  & 0.0244 & 377058 & 0.0554 & 221385 \\
$64e3$ & 0.0031 & 2533550 & 0.0093 & 1194665 & 0.0158 & 678000 & 0.0388 & 396305 \\
\bottomrule
\end{tabular}%
}
\caption{FFNO results across rollout lengths for Gray-Scott equation for Fig.\ref{fig:difficulty}.}
\label{tab:result_rollout}
\end{table*}

\clearpage
\begingroup
\setlength{\intextsep}{6pt}
\setlength{\floatsep}{8pt}
\setlength{\textfloatsep}{8pt}
\makeatletter
\setlength{\@fptop}{0pt}
\setlength{\@fpsep}{8pt}
\makeatother
\begin{table*}[!ht]
\centering
\small
\renewcommand{\arraystretch}{1.10}
\setlength{\tabcolsep}{6pt}
\resizebox{0.74\textwidth}{!}{%
\begin{tabular}{c|cc|cc|cc}
\toprule
\multirow{2}{*}{\textbf{Budget}}
& \multicolumn{2}{c|}{\textbf{$Re \in (10,40]$}}
& \multicolumn{2}{c|}{\textbf{$Re \in (40,90]$}}
& \multicolumn{2}{c}{\textbf{$Re \in (90,160]$}} \\
\cline{2-7}
& $\mathrm{nRMSE}_{\mathrm{avg}}$ & $N^\star_{\mathrm{avg}}$
& $\mathrm{nRMSE}_{\mathrm{avg}}$ & $N^\star_{\mathrm{avg}}$
& $\mathrm{nRMSE}_{\mathrm{avg}}$ & $N^\star_{\mathrm{avg}}$ \\
\hline
16e4 & 0.0088 & 4965 & 0.0820 & 2416 & 0.2897 & 956  \\
25e4 & 0.0077 & 7720 & 0.0602 & 2871 & 0.2892 & 1493 \\
33e4 & 0.0071 & 9874 & 0.0497 & 3707 & 0.2889 & 1972 \\
42e4 & 0.0043 & 8783 & 0.0386 & 4242 & 0.2840 & 2510 \\
\bottomrule
\end{tabular}%
}
\caption{FFNO results across Reynolds bins on BreakFlow for Fig.~\ref{fig:difficulty}.}
\label{tab:result_bin}
\end{table*}

\begin{table*}[!ht]
\centering
\small
\renewcommand{\arraystretch}{1.10}
\setlength{\tabcolsep}{5pt}
\resizebox{0.92\textwidth}{!}{%
\begin{tabular}{c|cc|cc|cc}
\toprule
\multirow{2}{*}{\textbf{Budget}}
& \multicolumn{2}{c|}{\textbf{1D Kuramoto--Sivashinsky}}
& \multicolumn{2}{c|}{\textbf{2D Kuramoto--Sivashinsky}}
& \multicolumn{2}{c}{\textbf{3D Kuramoto--Sivashinsky}} \\
\cline{2-7}
& $\mathrm{nRMSE}_{\mathrm{avg}}$ & $N^\star_{\mathrm{avg}}$
& $\mathrm{nRMSE}_{\mathrm{avg}}$ & $N^\star_{\mathrm{avg}}$
& $\mathrm{nRMSE}_{\mathrm{avg}}$ & $N^\star_{\mathrm{avg}}$ \\
\hline
$4e3$  & 0.0025 & INF & 0.0107 & 188906  & 0.3026 & 6070  \\
$16e3$ & 0.0009 & INF & 0.0073 & 726583  & 0.0937 & 24281 \\
$64e3$ & 0.0002 & INF & 0.0035 & 2810154 & 0.0432 & 97061 \\
\bottomrule
\end{tabular}%
}
\caption{Poseidon-L results across PDE dimensionality for the Kuramoto--Sivashinsky equation.}
\label{tab:poseidon_dim}
\end{table*}

\begin{table*}[!ht]
\centering
\small
\renewcommand{\arraystretch}{1.10}
\setlength{\tabcolsep}{5pt}
\resizebox{\textwidth}{!}{%
\begin{tabular}{c|cc|cc|cc|cc}
\toprule
\multirow{2}{*}{\textbf{Budget}}
& \multicolumn{2}{c|}{\textbf{Rollout length = 25}}
& \multicolumn{2}{c|}{\textbf{Rollout length = 50}}
& \multicolumn{2}{c|}{\textbf{Rollout length = 100}}
& \multicolumn{2}{c}{\textbf{Rollout length = 200}} \\
\cline{2-9}
& $\mathrm{nRMSE}_{\mathrm{avg}}$ & $N^\star_{\mathrm{avg}}$
& $\mathrm{nRMSE}_{\mathrm{avg}}$ & $N^\star_{\mathrm{avg}}$
& $\mathrm{nRMSE}_{\mathrm{avg}}$ & $N^\star_{\mathrm{avg}}$
& $\mathrm{nRMSE}_{\mathrm{avg}}$ & $N^\star_{\mathrm{avg}}$ \\
\hline
$4e3$
& 0.0062 & 71186
& 0.0066 & 50603
& 0.0068 & 17286
& 0.0080 & 11570 \\

$16e3$
& 0.0021 & 284742
& 0.0023 & 202411
& 0.0027 & 69145
& 0.0031 & 43181 \\

$64e3$
& 0.0013 & 1138970
& 0.0013 & 714297
& 0.0016 & 276579
& 0.0018 & 137674 \\
\bottomrule
\end{tabular}%
}
\caption{Poseidon-L results across rollout lengths for the Gray--Scott equation.}
\label{tab:poseidon_rollout}
\end{table*}

\begin{table*}[!ht]
\centering
\small
\renewcommand{\arraystretch}{1.10}
\setlength{\tabcolsep}{6pt}
\resizebox{0.74\textwidth}{!}{%
\begin{tabular}{c|cc|cc|cc}
\toprule
\multirow{2}{*}{\textbf{Budget}}
& \multicolumn{2}{c|}{\textbf{$Re \in (10,40]$}}
& \multicolumn{2}{c|}{\textbf{$Re \in (40,90]$}}
& \multicolumn{2}{c}{\textbf{$Re \in (90,160]$}} \\
\cline{2-7}
& $\mathrm{nRMSE}_{\mathrm{avg}}$ & $N^\star_{\mathrm{avg}}$
& $\mathrm{nRMSE}_{\mathrm{avg}}$ & $N^\star_{\mathrm{avg}}$
& $\mathrm{nRMSE}_{\mathrm{avg}}$ & $N^\star_{\mathrm{avg}}$ \\
\hline
$16e4$
& 0.0143 & 5371
& 0.0311 & 1656
& 0.1854 & 971 \\

$33e4$
& 0.0091 & 11078
& 0.0261 & 3416
& 0.1452 & 2002 \\

$42e4$
& 0.0064 & 12094
& 0.0231 & 4347
& 0.1122 & 2548 \\
\bottomrule
\end{tabular}%
}
\caption{Poseidon-L results across Reynolds bins on BreakFlow.}
\label{tab:poseidon_bin}
\end{table*}
\endgroup

\begingroup
\setlength{\intextsep}{6pt}
\section{Training-seed variability}
\label{Adx:seed_variability}
We evaluate FFNO and Poseidon-T across all four benchmarks and training budgets using three distinct training seeds while keeping the evaluation set fixed. Table~\ref{tab:seed_all} reports the mean and standard deviation of nRMSE across seeds, along with the corresponding breakeven complexity.

We observe that $N^\star$ remains unchanged across seeds for every setting. Although predictive errors vary between runs, these variations do not affect the matched classical solver configuration. With a fixed inference cost, this leads to identical breakeven complexities across seeds, suggesting that our estimates are robust to training randomness.

\begin{table}[!ht]
\centering
\small
\begin{tabular}{llrcr}
\toprule
Dataset & Model & Budget $B$ (s) & nRMSE ($\pm$ SD) & $N^\star_{\mathrm{avg}}$ \\
\midrule
\multirow{14}{*}{Navier--Stokes} & \multirow{7}{*}{FFNO} & 1,000 & $0.0070 \pm 0.000933$ & 32,158 \\
 &  & 2,000 & $0.0058 \pm 0.000344$ & 57,689 \\
 &  & 4,000 & $0.0041 \pm 0.000221$ & 101,796 \\
 &  & 8,000 & $0.0034 \pm 0.000038$ & 203,692 \\
 &  & 16,000 & $0.0025 \pm 0.000036$ & 407,383 \\
 &  & 32,000 & $0.0020 \pm 0.000034$ & 814,768 \\
 &  & 64,000 & $0.0013 \pm 0.000030$ & 1,629,536 \\
\cmidrule(lr){2-5}
 & \multirow{7}{*}{Poseidon-T} & 1,000 & $0.0060 \pm 0.000664$ & 27,918 \\
 &  & 2,000 & $0.0031 \pm 0.000385$ & 49,450 \\
 &  & 4,000 & $0.0014 \pm 0.000359$ & 93,240 \\
 &  & 8,000 & $0.0008 \pm 0.000098$ & 186,481 \\
 &  & 16,000 & $0.0003 \pm 0.000075$ & 372,962 \\
 &  & 32,000 & $0.0001 \pm 0.000075$ & 745,924 \\
 &  & 64,000 & $0.0001 \pm 0.000068$ & 1,491,848 \\
\midrule
\multirow{14}{*}{2D Kuramoto--Sivashinsky} & \multirow{7}{*}{FFNO} & 1,000 & $0.1506 \pm 0.005157$ & 533,930 \\
 &  & 2,000 & $0.0671 \pm 0.005114$ & 903,789 \\
 &  & 4,000 & $0.0331 \pm 0.001614$ & 598,991 \\
 &  & 8,000 & $0.0141 \pm 0.000469$ & 1,197,982 \\
 &  & 16,000 & $0.0067 \pm 0.000152$ & 2,243,547 \\
 &  & 32,000 & $0.0030 \pm 0.000124$ & 3,935,282 \\
 &  & 64,000 & $0.0019 \pm 0.000097$ & 7,346,417 \\
\cmidrule(lr){2-5}
 & \multirow{7}{*}{Poseidon-T} & 1,000 & $0.0190 \pm 0.001330$ & 45,862 \\
 &  & 2,000 & $0.0123 \pm 0.000247$ & 87,702 \\
 &  & 4,000 & $0.0069 \pm 0.000232$ & 175,405 \\
 &  & 8,000 & $0.0049 \pm 0.000038$ & 343,967 \\
 &  & 16,000 & $0.0037 \pm 0.000712$ & 687,934 \\
 &  & 32,000 & $0.0024 \pm 0.000063$ & 1,342,382 \\
 &  & 64,000 & $0.0016 \pm 0.000057$ & 2,684,765 \\
\midrule
\multirow{14}{*}{Gray--Scott} & \multirow{7}{*}{FFNO} & 1,000 & $0.3662 \pm 0.017652$ & 34,271 \\
 &  & 2,000 & $0.2296 \pm 0.015730$ & 35,571 \\
 &  & 4,000 & $0.1638 \pm 0.005966$ & 46,262 \\
 &  & 8,000 & $0.1165 \pm 0.004931$ & 92,524 \\
 &  & 16,000 & $0.0788 \pm 0.005391$ & 138,107 \\
 &  & 32,000 & $0.0535 \pm 0.003780$ & 221,385 \\
 &  & 64,000 & $0.0389 \pm 0.001037$ & 396,305 \\
\cmidrule(lr){2-5}
 & \multirow{7}{*}{Poseidon-T} & 1,000 & $0.0093 \pm 0.002250$ & 2,614 \\
 &  & 2,000 & $0.0053 \pm 0.000417$ & 5,294 \\
 &  & 4,000 & $0.0043 \pm 0.000292$ & 10,850 \\
 &  & 8,000 & $0.0036 \pm 0.000204$ & 22,178 \\
 &  & 16,000 & $0.0025 \pm 0.000105$ & 41,900 \\
 &  & 32,000 & $0.0017 \pm 0.000066$ & 67,869 \\
 &  & 64,000 & $0.0015 \pm 0.000120$ & 135,739 \\
\midrule
\multirow{8}{*}{BreakFlow} & \multirow{4}{*}{FFNO} & 160,000 & $0.1445 \pm 0.006902$ & 1,991 \\
 &  & 250,000 & $0.1418 \pm 0.009156$ & 2,139 \\
 &  & 330,000 & $0.1334 \pm 0.009034$ & 3,096 \\
 &  & 420,000 & $0.1189 \pm 0.004910$ & 3,068 \\
\cmidrule(lr){2-5}
 & \multirow{4}{*}{Poseidon-T} & 160,000 & $0.0821 \pm 0.005060$ & 1,327 \\
 &  & 250,000 & $0.0795 \pm 0.005840$ & 1,826 \\
 &  & 330,000 & $0.0745 \pm 0.004545$ & 2,737 \\
 &  & 420,000 & $0.0715 \pm 0.001633$ & 3,483 \\
\bottomrule \\
\end{tabular}
\caption{Three-seed robustness results across four PDE benchmarks for FFNO and Poseidon-T. Despite variations in nRMSE across training seeds, all three seeds match the same classical solver configuration at each budget, resulting in identical breakeven complexities. This demonstrates the robustness of breakeven complexity to training randomness.}
\label{tab:seed_all}
\end{table}

\endgroup

\end{document}